\documentclass[final,3p,times]{elsarticle}
\usepackage{amssymb}
\usepackage{algorithm}
\usepackage{algpseudocode}
\usepackage{amsmath}
\usepackage{xurl}
\usepackage{booktabs}
\usepackage{xcolor}
\usepackage{algpseudocode}
\usepackage{etoolbox}

\usepackage[english]{babel}
\usepackage{amsthm}

\usepackage{gensymb}
\usepackage{multirow}

\journal{Computer Methods in Applied Mechanics and Engineering}

\begin{document}

\begin{frontmatter}

\title{GALDS: A Graph-Autoencoder-based Latent Dynamics Surrogate model to predict neurite material transport}

\author[a]{Tsung Yeh Hsieh}
\author[a,b,c]{Yongjie Jessica Zhang}

\address[a]{Department of Mechanical Engineering, Carnegie Mellon University, 5000 Forbes Ave, Pittsburgh, PA 15213, USA}
\address[b]{Department of Biomedical Engineering, Carnegie Mellon University, 5000 Forbes Ave, Pittsburgh, PA 15213, USA}
\address[c]{Department of Civil and Environmental Engineering, Carnegie Mellon University, 5000 Forbes Ave, Pittsburgh, PA 15213, USA}

\begin{abstract}
Neurons exhibit intricate geometries within their neurite networks, which play a crucial role in processes such as signaling and nutrient transport. Accurate simulation of material transport in the networks is essential for understanding these biological phenomena but poses significant computational challenges because of the complex tree-like structures involved. Traditional approaches are time-intensive and resource-demanding, yet the inherent properties of neuron trees, which consists primarily of pipes with steady-state parabolic velocity profiles and bifurcations, provide opportunities for computational optimization. To address these challenges, we propose a Graph-Autoencoder-based Latent Dynamics Surrogate (GALDS) model, which is specifically designed to streamline the simulation of material transport in neural trees. GALDS employs a graph autoencoder to encode latent representations of the network's geometry, velocity fields, and concentration profiles. These latent space representations are then assembled into a global graph, which is subsequently used to predict system dynamics in the latent space via a trained graph latent space system dynamic model, inspired by the Neural Ordinary Differential Equations (Neural ODEs) concept. The integration of an autoencoder allows for the use of smaller graph neural network models with reduced training data requirements. Furthermore, the Neural ODE component effectively mitigates the issue of error accumulation commonly encountered in recurrent neural networks. The effectiveness of the GALDS model is demonstrated through results on eight unseen geometries and four abnormal transport examples, where our approach achieves mean relative error of \(3\%\) with maximum relative error \(<8\%\) and demonstrates a 10-fold speed improvement compared to previous surrogate model approaches.
\end{abstract}


\end{frontmatter}

\section{Introduction}
Material transport within neurite trees plays a critical role in maintaining neural function, supporting intracellular communication, and enabling the proper development and maintenance of neural networks. Disruptions to these transport processes are often associated with neurodegenerative diseases such as Alzheimer's and Parkinson's disease, where abnormalities in the structure of the neurite have been widely observed \cite{de2008role, gunawardena2005polyglutamine, kononenko2017retrograde, zhang2018modulation}. Accurate and efficient modeling of material transport in complex neurite geometries is essential to advance our understanding of neural health, disease progression, and potential therapeutic interventions.

To better understand material transport, researchers have proposed several mathematical models rooted in partial differential equations (PDEs). For instance, a foundational model described molecular motor-assisted transport of cell organelles and vesicles along filaments, which was subsequently extended to an axonal transport model incorporating the binding of multiple motor proteins \cite{smith2001models, friedman2005model}. While these models offer physically reasonable mechanistic explanations of transport in neurites, most initial studies simplified the complex neurite geometry by solving them in only one dimension (1D), ignoring the effects brought by the geometry \cite{wan2002one}. More recently, numerical techniques like the finite element method (FEM) have enabled simulations of these PDE models in more intricate geometries, yielding richer insights into cellular processes \cite{zhang2018geometric, zhang2013challenges}. For example, an FEM approach modeling extracellular electrical neural microstimulation revealed distinct neuron responses compared to hybrid FEM-cable equation methods \cite{loew2001virtual}. Building upon conventional FEM, isogeometric analysis (IGA) \cite{hughes2005isogeometric} was introduced to integrate geometric modeling with numerical simulation, offering enhanced accuracy and robustness,  and was applied in many biology modeling applications including cardiovascular modeling \cite{zhang2007patient, zhang2012atlas, urick2019review, yu2020anatomically} and neural growth simulation \cite{qian2022modeling, qian2025neurodevelopmental, qian20253d}. Our prior work involved developing an IGA solver to tackle 3D motor-assisted transport models and abnormal microtubule-induced traffic jam problems within diverse complex neurite networks \cite{li2019isogeometric, li2022modeling_sr, li2022modeling_jm}. 

However, the power of high-fidelity simulations comes at a significant price: prohibitive computational cost. A single, detailed simulation run can demand substantial computational resources, consuming hours, days, or even weeks on high-performance computing clusters. This computational burden represents a critical bottleneck, severely limiting the feasibility of analysis that require numerous simulation evaluations \cite{lu2019efficient}. Tasks such as extensive design space exploration, robust uncertainty quantification, inverse modeling, and iterative optimization often necessitate thousands or millions of simulation runs, rendering them impractical or economically infeasible with direct high-fidelity models. The challenges are compound by the fact that the complexity and computational demands of scientific models are continually increasing, driven by the desire of higher resolution, multi-physics coupling, and greater predictive accuracy. This escalation in simulation cost makes reliance solely on direct simulation increasingly unsustainable for many practical applications.

To alleviate this burden, the field has turned to surrogate modeling. A surrogate model is a computationally inexpensive approximation designed to mimic the input-output relationship of the original, costly simulation model \cite{azarhoosh2025review}. The primary objective of surrogate modeling is to drastically reduce the computational cost associated with evaluating the model's response, often by several orders of magnitude. Once trained, the surrogate model can be evaluated rapidly, enabling large-scale computational campaigns, such as extensive parameter sweeps for sensitivity analysis, optimization, or uncertainty quantification, that were previously intractable.

Deep learning has emerged as a powerful tool for constructing surrogate models to approximate complex physical simulations. Traditional fully connected deep neural networks (DNNs) have been applied to parametric PDEs and regression tasks in scientific machine learning \cite{raissi2017machine, tripathy2018deep}. While DNNs can fit low-dimensional mappings, they often fail to scale to high-dimensional input-output relationships and struggle with generalization due to the lack of inductive biases related to geometry or physics. To address spatial dependencies, Convolutional Neural Networks (CNNs), particularly encoder-decoder architectures like U-Net \cite{ronneberger2015u}, exploit spatial locality and have been used to model grid-based physical systems such as fluid flows \cite{thuerey2020deep}, heat conduction \cite{guo2016convolutional} and phase fields \cite{qian2023biomimetic}. While CNNs leverage spatial locality to improve learning efficiency, their reliance on regular grids limits their generalizabilty to irregular or dynamic meshes. Moreover, CNN-based surrogates require a large volume of data and computational resources, and suffer from difficulties when modeling long-term temporal dependencies or mutli-scale interactions.

For learning over non-Euclidean domains, Graph Neural Networks (GNNs) provide a natural framework, making them well-suited for surrogate modeling in mesh-based or particle-based simulations. GNN-based surrogates have shown promising results in learning dynamics across varying geometries \cite{pfaff2020learning, sanchez2020learning}, and have been successfully applied to simulate physics on unstructured meshes. However, their performance often degrades as the number of message-passing steps increases, and their expressiveness may be insufficient for capturing complex long-range dependencies without architecture modifications or hierarchical pooling schemes. Another significant development, Physics-Informed Neural Networks (PINNs), aims to incorporate known physical laws directly into the training process by penalizing the residual of governing PDEs \cite{raissi2019physics, hsieh2024multiscale}. This approach improves data efficiency and interpretabilty but typically requires retraining for each change in geometry, boundary condition, or parameter set, limiting scalability in multi-query contexts such as optimization or uncertainty quantification \cite{wang2021understanding}. Furthermore, PINNs can struggle with convergence and accuracy in stiff or advection-dominated systems \cite{krishnapriyan2021characterizing}.

More recently, transformer-based models \cite{vaswani2017attention}, which originated in natural language processing, have been adapted to physical modeling by exploiting their ability to model long-range dependencies \cite{qian2025high}. Architectures such as the graph transformer \cite{feng2024novel} and physics-aware attention mechanisms have shown potential in modeling complex spatiotemporal fields \cite{zhongbo2024pre, hemmasian2024multi}. However, standard transformers scale quadratically with the input size, making them computationally expensive for high-resolution PDE domains unless approximations like sparse attention or learned positional embeddings are introduced. A distinct class of methods, neural operators, are designed to learn mappings between function spaces, enabling surrogate models that generalize across different input fields. DeepONet \cite{lu2021learning} uses a branch-trunk architecture to learn operators directly from data and has demonstrated strong performance across parametric PDEs and many related research has been proposed \cite{li2023phase, zhu2023fourier, he2023novel}. Fourier Neural Operators (FNOs) \cite{li2020fourier, lehmann20243d, ye2024locality} further accelerate learning by performing convolution in Fourier space, allowing resolution-independent generalization. Despite their success, these methods still require training on full-field outputs, resulting in high memory and computational costs, and often exhibit limitations in capturing high-frequency or localized features.

In summary, existing deep learning surrogate models are often hampered by intensive data requirements, significant computational costs, difficulties in handing high-dimensional or geometrically complex data, and challenges in ensuring physical consistency and robust training. For example, in our previous studies, an IGA-based physics-informed graph neural network (PGNN) was developed to learn material transport mechanisms from simulation data, providing fast prediction within complex neurite networks \cite{li2021deep, li2023isogeometric}. While this work demonstrated significant efficiency improvements over conventional methods, the model's need to directly predict field values on the entire graph resulted in large number of GNN model parameters and substantial training data requirements, making the time for training and dataset generation still significant.

To overcome these pervasive challenges, particularly concerning scalability, generalization to irregular domains, and data inefficiency in material transport simulations, we introduce a new Graph-Autoencoder-based Latent Dynamics Surrogate (GALDS) model. Specifically engineered for simulating material transport within complex neurite trees, GALDS employs a graph autoencoder to effectively compress high-dimensional simulation outputs (e.g., concentration or velocity fields across cross-sections) into a compact, low-dimensional latent space. Within this latent space, the dynamics are robustly modeled using a Neural Ordinary Differential Equations (Neural ODEs) framework integrated with a GNN, enabling efficient temporal evolution of the system. This latent-space operation inherently reduces the number of learnable parameters and the requisite training data, while the graph representation adeptly captures the intricate topological and geometric complexities inherent to neurite structures. By synergistically combining graph-based learning, latent space modeling and continuous-time dynamics, GALDS presents a highly data-efficient and scalable alternative to conventional surrogate methods. The main contributions of this work are as follows:
\begin{itemize}
    \item Development of a novel graph autoencoder-based surrogate model framework for rapid and accurate prediction of material transport and traffic jams, caused by the reduction of molecular motors on microtubules, within complex 3D neurite geometries. 
    
    \item Introduction of a latent-space learning paradigm for dynamic system prediction, a departure from physical-space learning as seen in methods like PGNN. This approach significantly reduces both model complexity and the required training dataset size for predicting transport phenomena. 

    \item A novel integration of GNNs with a Neural ODEs framework for dynamic system prediction, moving beyond recurrent GNNs. This design learns continuous-time dynamics, enabling the use of numerical integration to obtain solutions at each timestep, thereby minimizing the error accumulation typically associated with long-term dynamic system predictions. To our knowledge, this represents the first combination of GNNs with Neural ODEs for predicting physical system dynamics. 

    \item Comprehensive demonstration of GALDS's superior performance compared to the PGNN method, exhibiting enhanced accuracy alongside improved computational and training data efficiency. 
    
\end{itemize}

The remainder of this paper is organized as follows. Section \ref{ch2} introduces the formulation of material transport and details the IGA solver employed for data generation and obtaining exact solutions. Section \ref{ch3} provides a concise overview of the proposed surrogate model concept and highlights the advantages of its development. A comprehensive and detailed discussion of the GALDS method's specifics is presented in Section \ref{ch4}. Section \ref{ch5} demonstrates the effectiveness of GALDS under both geometric and parametric complexities. In Section \ref{ch5.1}, for normal transport,  GALDS is evaluated on more complex geometries with symmetric flow patterns. In contrast, in \ref{ch5.2}, for traffic jams induced by spatially heterogeneous attachment/detachment kinetics, GALDS is tested under more complex parameter distributions and resulting in non-symmetric flow patterns. Finally, Section \ref{ch6} offers concluding remarks and outlines directions for future work.

\section{IGA based 3D simulation of Navier-Stokes equation and neural material transport} \label{ch2}
In previous studies, we have developed an IGA solver for dynamic material transport simulation in complex 3D neurite structures \cite{li2019isogeometric}. This IGA solver can provide accurate solution as the training data for the surrogate model. This section will first introduce the review and the advantage of the IGA method, and then the equivalents of the neural material transport will be shown.

A 1D model was developed to describe the macroscopic intracellular transport driven by the molecular motors \cite{smith2001models}. The equation is in a form of reaction-diffusion-transport which provides a description of the interaction between particles and filaments and also includes the free diffusion of unattached particles as well as directed motion of attached particles. Later, a more generalized form is derived to bring the simulation into more complicated 3D geometry for a single neurite \cite{li2019isogeometric}. We follow the same formulation:

\begin{equation} \label{Eq: Material Transport}
\left\{
\begin{aligned}
& \frac{\partial{n_{0}}}{\partial{t}} - D\nabla^2n_0 =  -(k_++k_-)n_0 + k_{+}^{'}n_+ + k_{-}^{'}n_- &\text{in }\Omega \\
& \frac{\partial{n_+}}{\partial{t}} + \boldsymbol{v}_+\cdot\nabla{n_+} =  k_+n_0-k_{+}^{'}n_+ &\text{in }\Omega\\
& \frac{\partial{n_-}}{\partial{t}} + \boldsymbol{v}_-\cdot\nabla{n_-} =  k_-n_0-k_{-}^{'}n_- &\text{in }\Omega\\
\end{aligned}
\right.
\end{equation} 

 \noindent where \(n_0\), \(n_+\) and \(n_-\) are the spatial concentrations of free, incoming and outgoing materiels, respectively. \(D\) represents the known diffusion coefficient; \(v_+\) and \(v_-\) are the velocities of incoming and outgoing material respectively. The two velocities will be solved separately and applied in Eq. (\ref{Eq: Material Transport}). The \(k_+\), \(k_-\), \(k_{+}^{'}\) and \(k_{-}^{'}\) are the cytoskeletal filament attachment rate of incoming and outgoing material, detachment rate of incoming and outgoing material, respectively.

We assume stable concentration of free and incoming materials at both the incoming and outgoing ends of the neurite as the boundary condition of solving Eq. (\ref{Eq: Material Transport}). The boundary conditions can be written as:

\begin{equation} \label{Eq: Material Transport BC}
    \left\{
    \begin{aligned}
        & n_0 = n_i, & n_+ = \lambda_i{n_i} &\quad \text{at } \Gamma_{in} \\
        & n_0 = n_o, & n_- = \lambda_o n_o &\quad \text{at } \Gamma_{out} \\
    \end{aligned}
    \right.
\end{equation}

\noindent where \(\Gamma_{in}\) and \(\Gamma_{out}\) are the boundaries at the incoming and outgoing ends of the neurite tree. The \(\lambda_i\) and \(\lambda_o\) control the filament attachment at both boundaries, which works like a loading. In this study, we assume the filament system is unipolar that leads to a unidirectional material transport process and ignore \(n_{-},\boldsymbol{v}_-, k_-,k'_-\) terms in Eq. (\ref{Eq: Material Transport})

The velocity profile in Eq. (\ref{Eq: Material Transport}) is solved in a separated solver. We first assume that the flow of the transport is incompressible and that the flow remains in a steady state then we solve the incompressible Navier-Stokes equation with an IGA solver, the diffusion coefficient of the flow is also assume constant. The incompressible Navier-stokes equation can be written as:
\begin{equation} \label{Eq: Navier-Stokes}
\left\{
\begin{aligned}
    & \nabla \cdot \boldsymbol{u} = 0 & \text{in }\Omega\\
    & \nabla \cdot (\boldsymbol{u} \otimes \boldsymbol{u}) + \nabla{\hat{p}} = \nu \nabla{\boldsymbol{u}} + f & \text{in }\Omega \\
\end{aligned}
\right.
\end{equation}

\noindent where \(\boldsymbol{u}\) and \(\hat{p}\) represent the flow velocity and pressure, respectively. \(f\) and \(\nu\) are the body force and kinematic viscosity. The no-slip boundary condition was applied to all the neurite wall and a parabolic velocity profile was applied to the inlet boundary as:

\begin{equation} \label{Eq: Navier-Stokes BC}
    \boldsymbol{u}(r) = u_t (1-(\frac{r}{R})^2) \boldsymbol{n},
\end{equation}

\noindent where \(u_t\) is the defined inlet transport velocity, \(\boldsymbol{n}\) is the normal vector of the inlet surface, \(r\) is the distance from the centerline to the evaluation point, and \(R\) is the radius of the cross section. The velocity result is then used in Eq. (\ref{Eq: Material Transport}) with \(\boldsymbol{v}_+ = \boldsymbol{u}\) and \(\boldsymbol{v}_- = -\boldsymbol{u}\). In addition, the streamline upwind Petrov-Galerkin (SUPG) formulation is implemented to enhance numerical stability. The velocity profile obtained is then substituted into Eq. (\ref{Eq: Material Transport}) to solve for the unknown concentrations.

IGA is employed to solve the equations, primarily because it offers a more precise geometric representation and improved accuracy when modeling the complex neurtie tree. A key advantage of IGA is its use of identical, smooth spline basis functions for both defining the geometry and performing the analysis. In this work, we utilize the truncated hierarchical B-splines (THB-splines) \cite{wei2017truncated, wei2017truncated2} for this purpose, which ensures high-order continuity and smoothness within the individual geometric elements of the computational domain. The blending functions are also defined using Bernstein-B\'ezier splines on unstructured meshes to handle extraordinary points (EPs) \cite{wei2018blended} in the neurite tree control meshes.

To explain the core concepts and formulation, we first revisit the standard B-spline basis functions, which form the foundation of the THB-spline construction. The mathematical formulation of B-splines is based on a piecewise polynomial construction defined by a knot vector \(\Xi=\{\xi_1, \xi_2, ..., \xi_{m+\bar{p}+1}\}\), where \(\xi_i\) is the \(i\)th knot, \(m\) is the number of basis functions and \(\bar{p}\) is the polynomial degree. The B-spline basis \(\{N_{i,\bar{p}}\}^{m}_{i=1}\)can then be defined with the Cox-de Boor recursion formula \cite{boor1971subroutine} as: 

\begin{equation} \label{Eq: B-spline basis p=0}
N_{i,0}(\xi) = \left \{
\begin{aligned}
    & 1 & \quad \text{if } \xi_i \leq x_i \leq \xi_{i+1} \\
    & 0 & \quad \text{otherwise} \\
\end{aligned}
\right.
\end{equation}

\noindent and for \(\bar{p}\geq1\) we have:
\begin{equation} \label{Eq: B-spline basis p>=1}
    N_{i,\bar{p}}(\xi) = \frac{\xi - \xi_i}{\xi_{i+\bar{p}}-\xi_i}N_{i,\bar{p}-1}(\xi)+\frac{\xi_{i+\bar{p}+1}-\xi}{\xi_{i+\bar{p}+1}-\xi_{i+1}}N_{i+1,\bar{p}-1}(\xi).
\end{equation}

\noindent These B-spline basis enable flexible and efficient representation of complex geometries and the solution space.

To facilitate integration with standard FEM solvers, B\'ezier extractions is used to express B-spline basis functions in terms of Bernstein polynomials, maintaining compatibility with element-wise assembly while preserving IGA's benefits.
We can express the domain by:
\begin{equation} \label{Eq: Surface Bezier representation}
    V(\xi, \eta) = (\boldsymbol{Q}^0)^T\boldsymbol{B}^0(\xi, \eta),
\end{equation}

\noindent where \(\boldsymbol{Q}^0\) are the B\'ezier control points and the associated B\'ezier functions \(\boldsymbol{B}^0\) are defined as:
\begin{equation} \label{Eq: Bezier function}
    \boldsymbol{B}^0(\xi, \eta)=[B^0_0(\xi, \eta), B^0_1(\xi, \eta), B^0_1(\xi, \eta)]^T,
\end{equation}
\noindent and
\begin{equation} \label{Eq: Bezier function component}
    B^0_0(\xi, \eta)=b_{i\%4}(\xi)b_{i/4}(\eta), \quad i=0,1,...,15.
\end{equation}
The \(\%\) and \(/\) represents ``modulo" and ``integer division" respectively. The functions \(b_j(t) ,\:\forall j \in(0,3)\) are univariate cubic Bernstein polynomials written as:
\begin{equation} \label{Eq: Bernstein polynomials}
    b_0(t)=(1-t)^3, \quad b_1(t)=3(1-t)^2t, \quad b_2(t)=3(1-t)t^2, \quad b_3(t)=t^3.
\end{equation}

THB-splines extend the B-spline framework by enabling local refinement through hierarchical structuring and truncation mechanisms. The B\'ezier extraction process,  applied at the element level, remains compatible with this framework and facilitates integration with standard finite element assembly pipelines. In our case, the generalized THB-spline framework with Bernstein-B\'ezier blending is employed to support unstructured hexahedral meshes and handle EPs. For each iterior element of the hexahedral control mesh, its eight control points are first used to define the body B\'ezier control points. Then, face, edge, and corner B\'ezier control points are computed by averaging the nearst body points, ensuring smooth transitions and continuity across elements. For boundary elements, a similar approach is used, except that the boundary B\'ezier control points are derived from the associated quadrilateral boundary mesh \cite{casquero2020seamless,wei2022analysis}. This unified procedure allows consistent construction of blending functions and B\'ezier extraction operators throughout the entire domain, enabling effective treatment of EPs and seamless integration into the overall spline framework.

The control mesh is first generated by the skeleton-based sweeping method \cite{zhang2007patient} which builds hexahedral control mesh from the given skeleton information. Then, the B\'ezier extraction is preformed as in Eq. (\ref{Eq: Surface Bezier representation}), allowing the weak forms of the Navier-Stokes equation Eq. (\ref{Eq: Navier-Stokes}) and the material transportation equation Eq. (\ref{Eq: Material Transport}) to be assembled using the B\'ezier control point \(\boldsymbol{Q}^0\) in the IGA solver. 

Due to the high geometric complexity of the neuron tree with more than one million control points, the calculation for such a big matrix  and the memory storage for intermediate result are highly demanding. Therefore the IGA solver is implemented with a parallelization library which separates the geometry into several parts and is run in parallel. The PETSc library is also used for parallelization and matrix solving, and the GMRES solver is employed with a proper restart setting to reduce memory demand. The solver is deployed in Pittsburgh Supercomputing Center (PSC) and the process is parallelized across nodes of PSC Bridges-2 CPUs.

\section{Overview of surrogate modeling for neurite transport simulation} \label {ch3}
\begin{figure}[htbp]
\centering
\setlength{\abovecaptionskip}{-5pt}
\includegraphics[width=\linewidth]{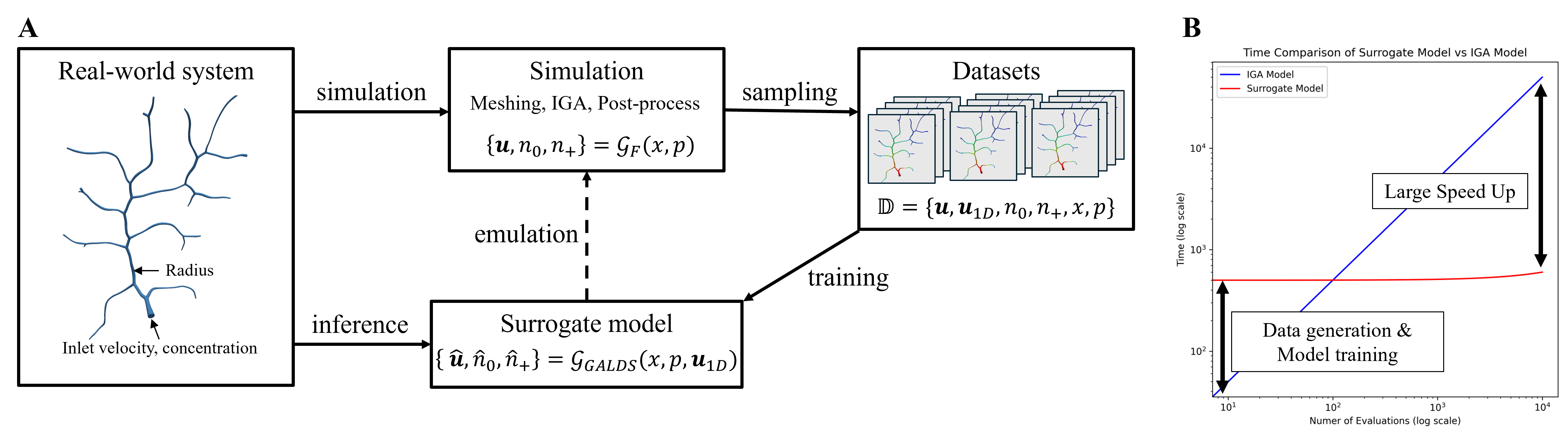}
\caption{(A) A schematic overview of a typical surrogate modeling workflow using neurite transport as an example.  (B) Time comparison of the surrogate model and IGA under parameter study that requires multiple simulation runs. Surrogate models have large time efficiency benefits when the number of evaluations is large.}
\label{figure: Framework Overview}
\end{figure}

This section overviews of our surrogate model for neurite transport simulations. Fig. \ref{figure: Framework Overview}(A) illustrates the overall pipeline for developing and applying a surrogate model to efficiently analyze complex systems. The process begins with defining a real-world system, capturing its essential physical and geometrical characteristics. This system is then translated into a computational domain, where high-fidelity numerical simulations are performed to generate detailed outputs. By running simulations across a range of input parameters, a dataset is constructed to represent the system's dynamic behavior. This dataset is then used to train a surrogate model, which is typically built using data-driven methods and learns to approximate the response of the full order simulation. Once trained, the surrogate model enables rapid inference, providing fast predictions for new input conditions.
 Fig. \ref{figure: Framework Overview}(B) compares the computational cost of performing multiple simulations using IGA versus the surrogate model.  Conventional methods typically require comparable computational time for each simulation run, leading to a total computational cost proportional to the number of simulations executed. In contrast, although surrogate modeling requires an initial phase of data generation and model training, it offers much lower cost for subsequent evaluations. As shown in the figure, for parameter studies involving many simulation queries, the surrogate model significantly reduces the overall computational effort compared to executing high fidelity IGA simulation repeatedly.

Our objective here is to elaborate on the development of our proposed surrogate model. The IGA simulation model, mathematically detailed in Section \ref{ch2} for the Navier-Stokes and material transport equations, provides the solution fields of interest, \(\boldsymbol{u}\) (velocity) and \(\{n_0, n_+\}\) (concentration), based on geometry \(x\) and parameters \(p=(D_{\pm}, k_{\pm}, l_{\pm})\). This can be represented by the system equation:
\begin{equation} \label{Eq: FEM model}
\{\boldsymbol{u}, n_0, n_+\}=\mathcal{G}_{F}(x, p).  
\end{equation}
We propose a surrogate model capable of predicting the solution field across the entire geometry domain. As discussed in Section \ref{ch2}, the full simulation is weakly coupled by solving the Navier-Stokes equation and the material transport equation. Although the velocity field exhibits a predominantly parabolic profile due to the low Reynolds number and can be rapidly approximated using a 1D reduced-order model (ROM) \cite{carson2017implicit}, this approximation neglects secondary flow effects introduced by geometric features such as curvature, tapering, and branching \cite{wan2002one}. These effects can lead to non-negligible deviations from the idealized profile and impact downstream transport predictions. 

However, the 1D ROM velocity result \(\boldsymbol{u}^h_{1D}\) can be very useful to reduce the surrogate model’s complexity by providing a low-dimensional, physics-informed representation of the velocity field as the input. This simplifies the learning task, as the model no longer needs to infer velocity patterns from raw geometry and parameters alone. Instead, a trained decoder refines the 1D ROM input to reconstruct the full spatial velocity field, implicitly capturing geometric influences beyond the 1D approximation. Simultaneously, the model predicts the concentration field \(n^h\)  by operating on a latent space derived from the 1D ROM data, integrating velocity and geometry in a compact representation.

Thus, the overall model jointly predicts both velocity and concentration fields from the 1D ROM input, combining physical insight with data-driven refinement. The model is trained on 3D simulation data and centerline-extracted features, enabling fast and accurate predictions from limited input information. The surrogate model can be concisely represented as follows:
\begin{equation} \label{Eq: Surrogate model}
    \{\hat{\boldsymbol{u}}, \hat{n}_0, \hat{n}_+\}=\mathcal{G}_{GALDS}(x, p, \boldsymbol{u}_{1D}),
\end{equation}
\noindent where \(\boldsymbol{u}_{1D}\) denotes the input information obtained from the 1D Navier-Stokes simulation. The GALDS model aims to provide a significantly more computationally efficient approach for parameter studies or obtaining simulation results from unseen neurite geometries.

The training data for the surrogate model is generated using the IGA solver, which simulates both the Navier-Stokes equation and the material transport equation in three dimensions. The generated results are then automatically segmented into sets corresponding to pipe and bifurcation components. Given the potentially large number of B\'ezier nodes in each simulation, we adopt an information extraction approach similar to that described in \cite{li2021deep}. Specifically, 17 nodes represent a single cross-section of a pipe, and 23 nodes represent a trifurcate template of a bifurcation. The distribution of these extracted nodes is further illustrated in Section \ref{ch4}.

\begin{figure}[htp]
\centering
\setlength{\abovecaptionskip}{-3pt}
\setlength{\belowcaptionskip}{-10pt}
\includegraphics[width=0.75\linewidth]{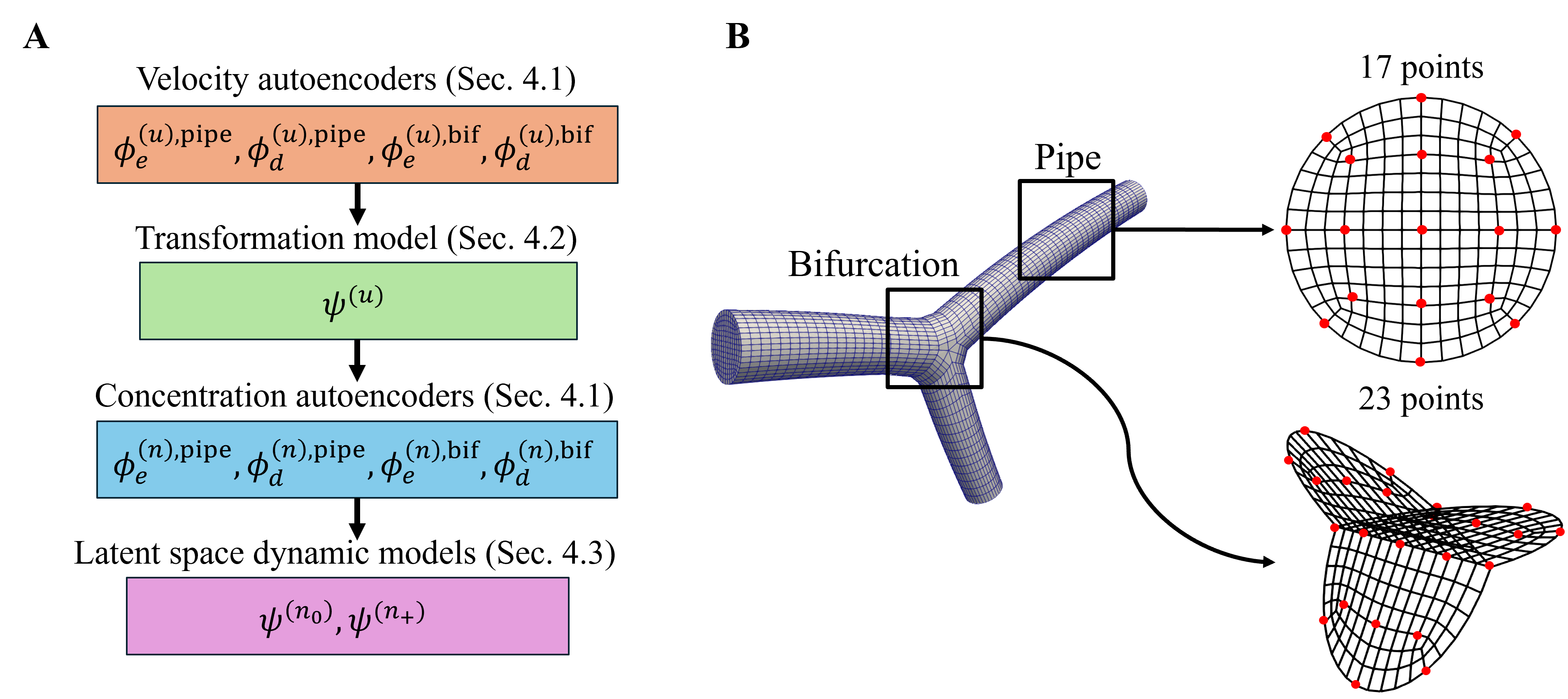}
\caption{(A) Training process of all GALDS modules. The velocity autoencoder utilizes separated models for pipe and bifurcation geometries, specifically \(\{\phi^{(n),\text{pipe}}_e,\phi^{(n),\text{pipe}}_d\}\) and \(\{\phi^{(n),\text{bif}}_e,\phi^{(n),\text{bif}}_d\}\) respectively, where the superscripts \(\circ^{(u),\text{pipe}}\) and \(\circ^{(u),\text{bif}}\) indicate the associated velocity field and geometry, and \(\{\circ_{e}, \circ_d\}\) denote the encoder and decoder components. A single latent space  transformation model \(\psi^{(u)}\) processes the shared velocity latent space. Similarly, the concentration autoencoder also employs separate models for pipe \(\{\phi^{(n),\text{pipe}}_e,\phi^{(n),\text{pipe}}_d\}\) and bifurcation geometries \(\{\phi^{(n),\text{bif}}_e,\phi^{(n),\text{bif}}_d\}\). The latent space system dynamic models \(\{\psi^{(n_0)}, \psi^{(n_+)}\}\) separates the prediction of concentrations  into \(n_0\) and \(n_+\). (B) Demonstration of pipe and bifurcation geometries with cross-sections. Red circles mark the extracted points on each cross-section.}
\label{figure: Training sequence & pipe bifurcation demo}
\end{figure}

\section{Graph-Autoencoder-based Latent Dynamics Surrogate model (GALDS)}  \label{ch4}
This section provides a detailed introduction to the architecture and mathematical underpinnings of the GALDS model. The GALDS framework comprises four distinct modules, and their sequential training process is outlined in Algorithm \ref{alg:training} in the Appendix and visually represented in Fig. \ref{figure: Training sequence & pipe bifurcation demo}(A). We have
 \begin{itemize}
     \item  Two pairs of velocity autoencoders (orange module), \(\{\phi_e^{(u),\text{pipe}}, \phi_d^{(u),\text{pipe}}\}\) and \(\{\phi_e^{(u),\text{bif}}, \phi_d^{(u),\text{bif}}\}\), trained to encode velocity data from the physical space into a low-dimensional latent space and to decode it back from the latent space to the physical space for pipe and bifurcation geometries;
     \item A latent space transformation model \(\psi^{(u)}\) (green module) designed to project physical centerline data directly into the latent space;
     \item  Two pairs of concentration autoencoders (blue module), \(\{\phi_e^{(n),\text{pipe}}, \phi_d^{(n),\text{pipe}}\}\) and \(\{\phi_e^{(n),\text{bif}}, \phi_d^{(n),\text{bif}}\}\), also trained for encoding and decoding concentration data between the physical and latent spaces for pipe and bifurcation geometries; and
     \item Two latent space system dynamics models \(\{\psi^{(n_0)}, \psi^{(n_+)}\}\) (purple module), responsible for modeling the temporal evolution within the latent space for two concentration fields.
 \end{itemize}

The GALDS architecture strategically incorporates the velocity and concentration autoencoders to specifically handle distinct geometric domains as depicted in Fig. \ref{figure: Training sequence & pipe bifurcation demo}(B): pipe (represented by 17 points per cross-section) and bifurcation structures (represented by 23 points per cross-section). This architectural design allows each model instance to specialize in the unique characteristics of its respective geometry. Concurrently, the latent space system dynamics module predicts two different concentration fields, \(n_0\) and \(n_+\), which are subsequently accumulated to form the final output. Crucially, this system dynamics model is uniquely structured to reconstruct information into a single, unified tree, facilitating prediction across the entire neurite tree without requiring separate processing of pipe and bifurcation segments.

The technical specifications and architectural details of each module are further discussed in subsequence dedicated sections. Specifically, Section \ref{ch4.1} will introduce the training process and detailed architecture of the velocity (orange module) and concentration (blue module) autoencoder models. Section \ref{ch4.2} will then describe the latent space transformation model (green module), followed by Section \ref{ch4.3}, which will present the latent space system dynamics model (pink module). Finally, Section \ref{ch4.4} will detail the complete inference procedure once all modules have been successfully trained.

\subsection{Autoencoders for nonlinear dimensional compression} \label{ch4.1}
An autoencoder is a specially designed neural network architecture for dimensionality reduction. An autoencoder consists of a paired structure, encoder and decoder. The encoder is denoted as \(\phi_e(\cdot;\theta_e):\mathbb{R}^{D_p}\rightarrow \mathbb{R}^{D_l}\) and is parameterized by \(\theta_e\). The decoder is denoted as \(\phi_d(\cdot;\theta_d):\mathbb{R}^{D_l}\rightarrow \mathbb{R}^{D_p}\) and parameterized by \(\theta_d\). The \(D_p\) and \(D_l\) are the physical and latent space dimensions, respectively, and generally they satisfy the relation \(D_p >> D_l\). The encoder conducts a nonlinear transformation which projects the input into the latent space that has less features, and the decoder projects the compressed feature space back to the physical dimension. The two models can be described by:

\begin{equation} \label{Eq: Auto-encoder}
\left \{
\begin{aligned}
& \tilde{\boldsymbol{z}} = \phi_e(\boldsymbol{X};\theta_e),\\
&\hat{\boldsymbol{X}} = \phi_d(\tilde{\boldsymbol{z}};\theta_d),\\
\end{aligned}
\right.
\end{equation}

\noindent where \(\boldsymbol{X}\in \mathbb{R}^{D_p}\) represents the sampling points of the physical domain input, \(\tilde{\boldsymbol{z}} \in \mathbb{R}^{D_l}\) is the compressed latent space result and \(\hat{\boldsymbol{X}}\in \mathbb{R}^{D_p}\) is the approximated physical domain result.

The loss function is defined as:
\begin{equation} \label{Eq: Auto-encoder loss function}
    \mathcal{L}_{AE}(\theta_e, \theta_d) = ||\hat{\boldsymbol{X}} - \boldsymbol{X}||^2_2 = ||\phi_d(\phi_e(\boldsymbol{X};\theta_e);\theta_d) - \boldsymbol{X}||^2_2,
\end{equation}
\noindent where the subscript \(\circ_{AE}\) denotes a general autoencoder reconstruction error, measuring how well the decoder output \(\hat{\boldsymbol{X}}\) matches the input \(\boldsymbol{X}\). Minimizing \(\mathcal{L}_{AE}\) ensures that the encoder-decoder pair learns an effective low-dimensional representation from which the original input can be accurately reconstructed. While typical autoencoders often feature symmetric architectures, where the encoder's fully connected layers progressively shrink and the decoder's expand, our research instead leverages a graph autoencoder that processes graph data for both input and output.

\begin{figure}[htbp]
\centering
\setlength{\abovecaptionskip}{5pt}
\setlength{\belowcaptionskip}{-10pt}
\includegraphics[width=1.0\linewidth]{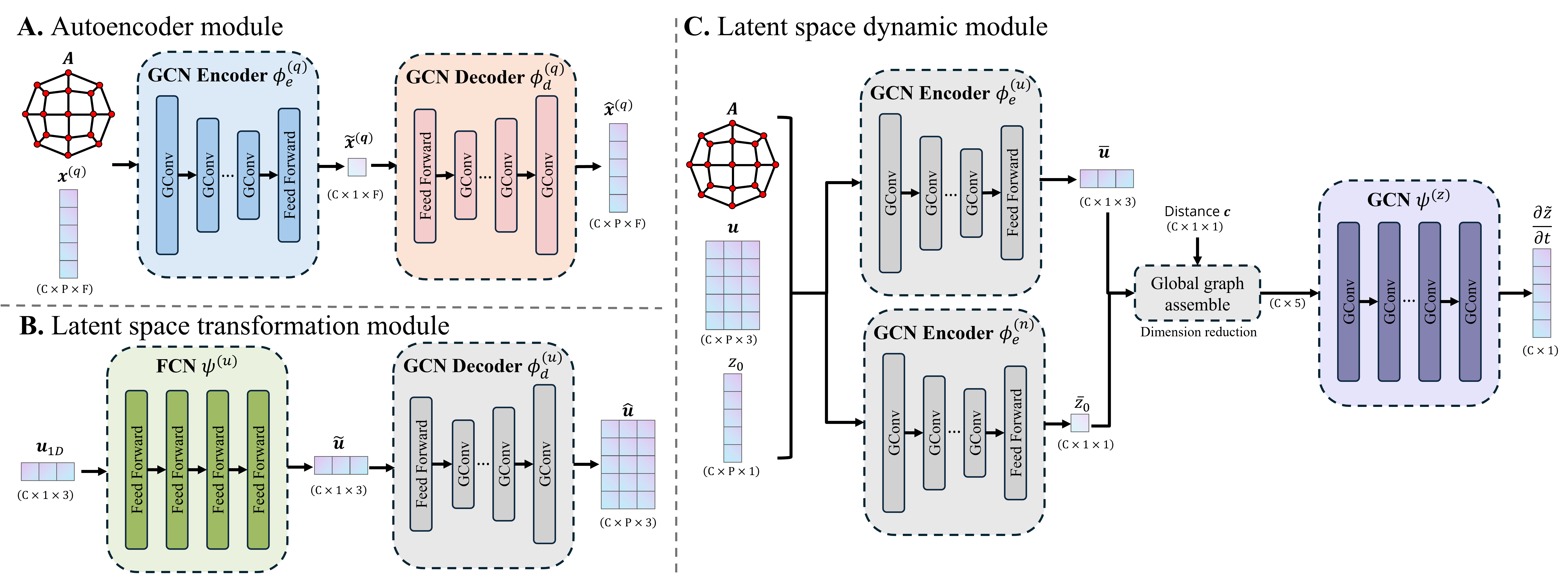}
\caption{Architecture of GALDS modules. (A) The autoencoder module takes the adjacency matrix \(A\) and node features \(\boldsymbol{x}^{(q)}\) as inputs, compresses them into the latent space representation \(\tilde{\boldsymbol{x}}^{(q)}\), and then reconstructs the original features \(\hat{\boldsymbol{x}}^{(q)}\). (B) The latent space transformation module uses a fully connected neural network (FCN) to map 1D reduced order solutions \(\boldsymbol{u}_{1D}\) to the latent space representation \(\tilde{\boldsymbol{u}}\), which is then decoded into physical space predictions \(\hat{u}\) using the decoder \(\phi_d^{(u)}\). (C) The latent space dynamic module uses the trained encoders \(\phi^{(u)}_e\) and \(\phi_e^{(n)}\) to obtain the latent velocity \(\bar{\boldsymbol{u}}\) and latent concentration \(\bar{z}\). A global graph assemble algorithm combines features across the entire neurite tree to construct a unified graph input for the model, which predicts the change in the temporal evolution of the latent concentration, \(\frac{\partial \tilde{z}}{\partial t}\).
The data dimensions are represented by capital letters, where \((C,P,F)\) denotes the number of cross-sections, the number of data points per cross-section, and the feature dimension, respectively. Blocks shown in gray are pre-trained with fixed parameters. ``GConv" and ``Feedforward" refer to individual graph convolution and fully connected layers, respectively. ``GCN" and ``FCN" represent network blocks composed of multiple GConv or feedforward layers. }
\label{figure: Sub-model details}
\end{figure}

As shown in Fig. \ref{figure: Sub-model details}(A), our framework employs two distinct autoencoder pairs to compress and reconstruct the simulation results: one for the 3D velocity fields (\(\boldsymbol{u}\)) from Navier-Stokes simulations and the other for the material transport concentrations (\(n_0\) and \(n_+\)). Each autoencoder consists of an encoder \(\phi_e^{(q)}\) and a decoder \(\phi_d^{(q)}\), where \(q\in \{u, n\}\)  is a label indicating whether the model operates on velocity (\(q=u\)) or concentration data (\(q=n\)). The input field is denoted by \(\boldsymbol{x}^{(q)}\), where \(\boldsymbol{x}^{(u)}=\boldsymbol{u}\) represents the velocity field and \(\boldsymbol{x}^{(n)}=(n_0,n_+)\) represents the concentration fields. The general form of the autoencoder is given by:
\begin{equation} \label{Eq: Auto-encoder model general}
\left \{
\begin{aligned}
    & \tilde{\boldsymbol{x}}^{(q)} = \phi_{e}^{({q})}(\boldsymbol{x}^{(q)}, \boldsymbol{A}; \theta_{e}^{({q})}),\\
    & \hat{\boldsymbol{x}}^{(q)} = \phi_{d}^{({q})}(\tilde{\boldsymbol{x}}^{(q)},\boldsymbol{A};\theta_{d}^{({q})}),\\
\end{aligned}
\right.
\end{equation}
\noindent where \(\tilde{\boldsymbol{x}}^{(q)}\) is its latent representation of the input, \(\hat{\boldsymbol{x}}^{(q)}\) is the reconstructed output, and \(\boldsymbol{A}\) is the adjacency matrix. The training objective for each autoencoder is to minimize the reconstruction error, defined as:

\begin{equation} \label{Eq: Auto-encoder model 1 loss function general}
    \mathcal{L}_{\phi^{({q})}}(\theta_{e}^{({q})}, \theta_{d}^{({q})}) = ||\hat{q}-q||^2_2  = ||\phi_{d}^{({q})}(\phi_{e}^{({q})}(\boldsymbol{x}^{(q)}, \boldsymbol{A}; \theta_{e}^{({q})}), \boldsymbol{A};\theta_{d}^{({q})}) - \boldsymbol{x}^{(q)}||^2_2.
\end{equation}

It is important to note that for a given target field (velocity or concentration), the autoencoders handling different geometries (pipe and bifurcation) share the same parameters for their encoders to maintain consistency in the latent space. These models are also trained simultaneously to optimize this consistency. By minimizing the two loss functions \(\mathcal{L}_{\phi^{({u})}}\) and \(\mathcal{L}_{\phi^{(n)}}\), we establish accurate nonlinear projections between the latent space and the physical space for both Naiver-Stokes (velocity) and material transport (concentration) simulations.

\subsection{Latent space transformation model for velocity prediction} \label{ch4.2}
A primary objective of our proposed GALDS model is to reduce reliance on extensive 3D simulation training data by incorporating embedded information from 1D ROM solutions. We achieve this by defining a projection, \(\psi^{(u)}\), that directly maps 1D simulation results from the physical space of the velocity autoencoder, \(\phi_{e}^{({u})}\). This projection effectively transfers insights from the computationally cheaper 1D model, guiding the higher-fidelity 3D reconstruction.

As illustrated in Fig. \ref{figure: Sub-model details}(B), the velocity in the latent space can be formally expressed as:

\begin{equation}
    \tilde{u}=\psi^{({u})}(\boldsymbol{u}_{1D};\theta_{\psi}^{(u)}),
\end{equation}

\noindent where \(\psi^{(u)}\) is the projection model parameterized by \(\theta_{\psi}^{(u)}\). Here,  \(\boldsymbol{u}_{1D}\) represents the velocity field obtained from the 1D Navier-Stokes simulation. In our implementation, \(\psi^{(u)}\) is realized as a fully connected neural network employing a ReLU activation function. The training of \(\psi^{(u)}\) aims to minimize the discrepancy between its output \(\tilde{\boldsymbol{u}}\) and the latent space representation generated by the pre-trained velocity encoder \(\phi_{e}^{(u)}\) with the loss function defined as:

\begin{equation}
    \begin{aligned}
        \mathcal{L}_{\psi^{(u)}} &=||\tilde{u} - \bar{u}||^2_2 = ||\psi^{(u)}(\boldsymbol{u}_{1D};\theta_{\psi}^{(u)}) - \phi_{e}^{(u)}(\boldsymbol{u}, \boldsymbol{A}; \theta_{e}^{(u)})||^2_2.
    \end{aligned}
\end{equation}

\noindent To ensure meaningful supervision, it is imperative that the velocity autoencoder \(\{\phi_{e}^{(u)}, \phi_{d}^{(u)}\}\) is well trained prior to training \(\psi^{(u)}\), as the loss function relies on the encoded latent vectors from \(\phi_{e}^{(u)}\). Notably, this training process utilizes the same dataset as that for the velocity autoencoder \(\{\phi_{e}^{(u)}, \phi_{d}^{(u)}\}\), preventing any additional computational overhead for data generation.

Upon successful training of \(\psi^{(u)}\), an accurate mapping from the 1D simulation results to the compressed latent spaces is established. This enables a streamlined pipeline to predict 3D Navier-Stokes velocity fields directly from the 1D simulation output, which can be represented as:
\begin{equation}
    \hat{u} = \phi_{d}^{(u)}(\psi^{(u)}(\boldsymbol{u}_{1D}),\boldsymbol{A}).
\end{equation}

\subsection{Graph Neural ODE for latent space system dynamic} \label{ch4.3}

In our proposed method, we utilize a neural network model to learn the system dynamics of material transportation. Rather than learning these system dynamics in a high-dimensional physical space, the model directly operates on the captured latent space data, which is discovered by the autoencoder models introduced in Section \ref{ch4.1}.  Since we aim to predict two concentrations, \(n_0\) and \(n_+\), we employ two separate, but identical, models for their respective predictions. For clarity, we will use \(z\) to generally represent either concentration in the subsequent derivations (\(z\in {\{n_0, n_+}\}\)). This latent space dynamics model, denoted as \(\psi^{(z)}\) and illustrated in Fig. \ref{figure: Sub-model details}(C), takes the following form:

\begin{equation}
\left \{
\begin{aligned}
    & \tilde{z}_t = \psi^{(z)}(\bar{z}_0, \bar{\boldsymbol{u}}; \theta_{\psi}^{(z)}),\\
    & \bar{z}_0 = \phi_{e}^{(n)}(z(t=0),\boldsymbol{A};\theta_e^{(n)}),
\end{aligned}
\right.
\end{equation}

\noindent where \( \tilde{z}_t\) is the predicted latent space result of the target concentration field at time \(t\), which can subsequently be decoded into the physical space solution using the concentration decoder \(\phi_{d}^{(n)}\). The latent velocity \(\bar{\boldsymbol{u}}\) is predicted by encoder \(\phi_e^{(u)}\). The term \( \bar{z}_0\) represents the latent space encoding of the initial condition and serves as an input to the model.  Following the approach of treating the velocity solution from Navier-Stokes as an input in IGA material transport simulations, we similarly utilize the latent space velocity \(\tilde{u}\) as an input to the model. Both latent space inputs (\(\bar{u},  \bar{z}_0\)) are assembled into a global graph, preserving the original mesh connectivity. This design allows the model to predict latent space results for the entire neurite tree in a unified graph format, avoiding segmentation into pipe and bifurcation sections during the prediction phase.

It is important to note that during the training phase, the two encoders \(\phi_{e}^{(u)}\) and \(\phi_{e}^{(n)}\) are ``frozen", meaning their parameters are not updated, and thus, related gradients are not computed as part of \(\psi^{(z)}\)'s training. This freezing mechanism is implemented to allow \(\psi^{(z)}\) to focus solely on learning the dynamic transport features, minimizing potential conflicts that could arise from simultaneous training of two interconnected models. Here, we employ a Graph Neural ODE model \cite{poli2019graph} as \(\psi^{(z)}\).

The Graph Neural ODE uses the similar idea as Neural ODEs \cite{chen2018neural}. Unlike traditional GNNs, which aim to predict state changes at discrete steps (e.g., \( z_{t+\Delta t} = z_t + \psi^{(z)}(z_t, \theta _{\psi}^{(z)})\) where \(z_t\) denotes the target field \(z\) at time \(t\) and \(t\in[0,T-\Delta t]\), Graph Neural ODEs directly model the time derivative. This is formulated as: 
 \begin{equation} \label{Eq: Nerual ODE}
  \frac{\partial \tilde{z}}{\partial t} = \psi^{(z)}(\bar{z}_0, \bar{\boldsymbol{u}},\boldsymbol{c}, t;\theta _{\psi}^{(z)}),
 \end{equation}
\noindent where \(\boldsymbol{c}\) are distance embeddings that represent the physical distance from each node to the root of the graph. This design helps mitigate error accumulation, a common issue in recurrent neural networks, by learning the continuous system behavior rather than discrete state transitions.

In our approach, the model  \(\psi^{(z)}\) is realized as a GNN comprising multiple graph convolutional layers, which renders it generalizable and applicable to any graph structure without model modification. We choose to use a sigmoid activation function, drawing inspiration from a work that also employs Graph Neural ODEs for classification, trajectory extrapolation, and traffic forecasting problems \cite{poli2019graph}. While our application involves predicting latent space derivatives, which is typically a regression problem, the sigmoid activation function demonstrated greater robustness during our internal testing.

Once the time derivative is predicted by the GNN, the actual latent space result \(z\)  can be obtained by applying the 4th-order Runge-Kutta (RK4) method or other standard numerical integration schemes,  a process we encapsulate with the notation \(\text{Int}(\circ)\). The training data for this model consists of IGA simulation results from the initial time step to the steady state. The loss function for model training is then calculated as:
\begin{equation} \label{Eq: Recurrent Model Loss Total}
\begin{aligned}
    \mathcal{L}_{\psi_2} = ||\tilde{z}(t_i +\Delta t) - \bar{z}(t_i +\Delta t)||_2^2= ||\text{Int}(\psi^{(z)}(\bar{z}(t_i),\bar{\boldsymbol{u}}, \boldsymbol{c}, t_i),\Delta t) - \bar{z}(t_i +\Delta t)||_2^2, \quad \text{for } t_i\in(0,T-\Delta t],
\end{aligned}
\end{equation}

\noindent where \(\bar{z}\) is derived by encoding the exact solution from IGA simulation using the concentration encoder:

\begin{equation} \label{Eq: Latent concentration decoding}
\begin{aligned}
    \bar{z} = \phi^{(n)}_e(n_0,\boldsymbol{A};\theta_e^{(n)}) \quad \text{or} \quad \bar{z} = \phi^{(n)}_e(n_+, \boldsymbol{A};\theta_e^{(n)}),
\end{aligned}
\end{equation}
depending on the target field being predicted.

The latent space system dynamics model for \(n_0\) and \(n_+\) are first trained separately until their respective losses converge. Subsequently, both models are joints fine-tuned with a new target field,  \(z=n_0+n_+\). Our internal testing indicates that this training technique yields the best accuracy.

\subsection{Model inference process} \label{ch4.4}
The model inference process commences immediately after the training phase, leveraging the fully trained neural network components to efficiently generate predictions from new, unseen inputs. A key prerequisite is that all trained neural network models have achieved a loss value below a predefined threshold, thereby ensuring their accuracy. Our specific objective is to accelerate the prediction of 3D Navier-Stokes simulation results (as in Eq. \ref{Eq: Navier-Stokes} and material transport simulation results (as in Eq. \ref{Eq: Material Transport}, with these predictions significantly enhance by the information derived from 1D Navier-Stokes simulations. Consequently, for a typical inference process, only two types of pre-generated data are required: the velocity filed from the 1D Navier-Stokes simulation (which includes velocity values along the centerline of the geometry) and 1D skeleton data (comprising centerline coordinates, cross-sectional diameters, and the connectivity relationships of the centerline points).

\begin{figure}[htp]
\centering
\setlength{\abovecaptionskip}{5pt}
\setlength{\belowcaptionskip}{-10pt}
\includegraphics[width=1.0\linewidth]{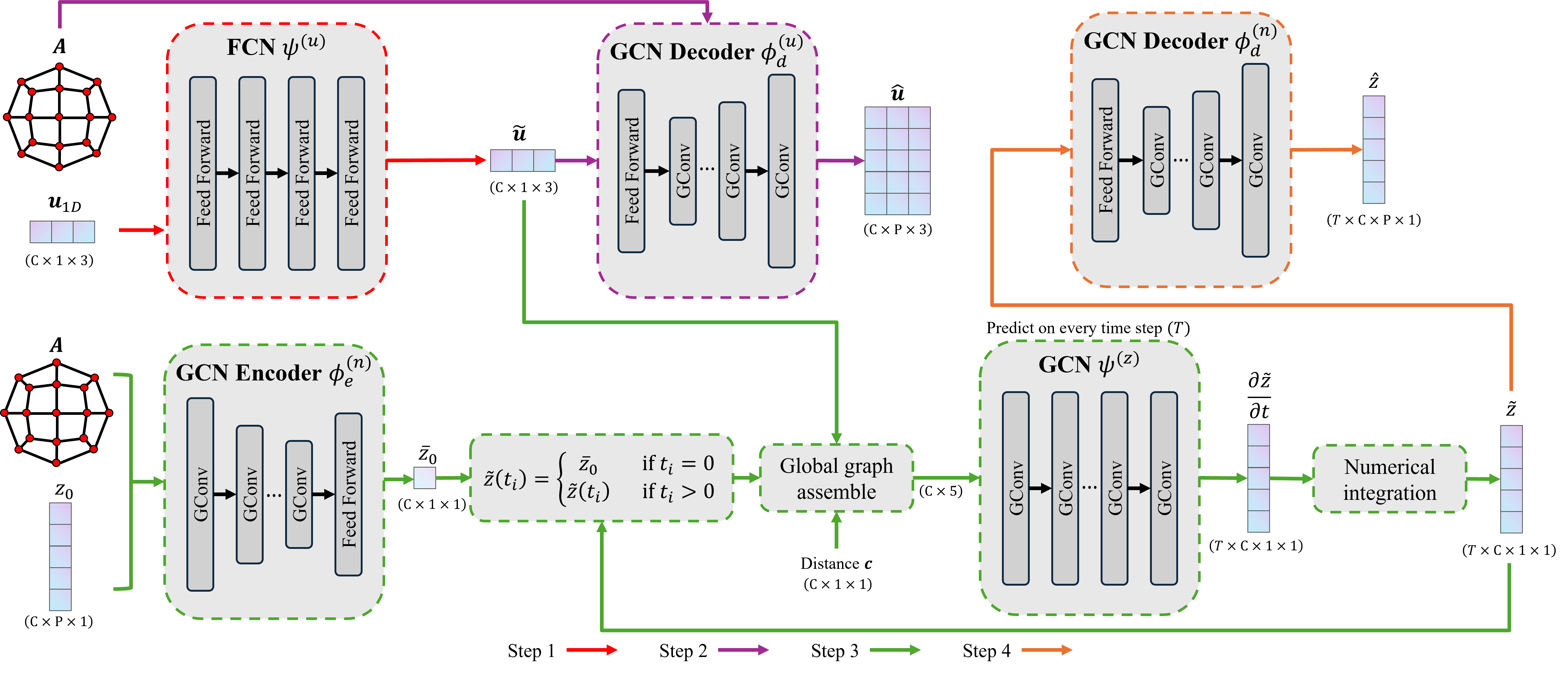}
\caption{Pipeline for the inference process of the GALDS model. This diagram outlines the four sequential steps, with each step clearly distinguished by a unique color (refer to the legend for details): red signifies the latent space projection of 1D velocity data, purple indicates the 3D velocity reconstruction, green represents the latent space concentration prediction over time, and orange denotes the final 3D concentration reconstruction. The data dimensions follow the same convention as presented in Fig. \ref{figure: Sub-model details}, and the additional dimension \(T\) is used to show the results in time steps.}
\label{fig:Model inference pipeline}
\end{figure}

As shown in Fig. \ref{fig:Model inference pipeline}, the model inference is systematically separated into the following four steps and demonstrated in four colors. The dimension of the input and output data of each block are also indicated; \((C,P,F)\) denotes the number of cross-sections, the number of data points per cross-section, and the feature dimension, respectively. For time-dependent data, an additional time dimension is added as \((T,C,P,F)\).
\begin{itemize}
    \item \textbf{Step 1: Latent Space Projection of 1D Velocity (Red).} The velocity data from the 1D Navier-Stokes simulation, \(\boldsymbol{u}_{1D}\in \mathbb{R}^{C\times1\times3}\), is fed into the latent space transformation model \(\psi^{(u)}\), which is a FCN block. This model outputs the corresponding latent space representation of the 1D simulation:
    \begin{equation}
        \tilde{\boldsymbol{u}}=\psi^{(u)}(\boldsymbol{u}_{1D}) \in \mathbb{R}^{C\times1\times3}.
    \end{equation}

    \item \textbf{Step 2: 3D Velocity Reconstruction (Purple).} The latent space representation \(\tilde{\boldsymbol{u}}\in\mathbb{R}^{C\times1\times3}\) and the adjacency matrix \(\boldsymbol{A}\) are then used as input for the velocity decoder \(\phi_{d}^{(u)}\). The decoder reconstructs the 3D velocity field in physical space, denoted as \(\hat{\boldsymbol{u}}\in\mathbb{R}^{C\times P \times 3}\).  We have:
\begin{equation}
        \hat{\boldsymbol{u}} = \phi_{d}^{(u)}(\tilde{\boldsymbol{u}},\boldsymbol{A}).
\end{equation}
\noindent The reconstructed \(\hat{\boldsymbol{u}}\) is then assembled into a semi-structured mesh file for visualization.
    \item \textbf{Step 3: Latent Space Concentration Prediction (Green).}  The process begins by encoding the initial physical concentration \(z_0\in \mathbb{R}^{C\times P \times 1}\). This initial physical concentration \(z_0\) and adjacency matrix \(\boldsymbol{A}\) are input into the GCN concentration encoder \(\phi^{(n)}_e\), which gives the initial latent space concentration \(\bar{z}_0\in\mathbb{R}^{C \times 1 \times 1}\). The initial latent space concentration then serves as the starting point, \(\tilde{z}(t_0)=\bar{z}_0\), for the recursive process managed by the Graph Neural ODE model.

At each time step of the prediction, the model first assembles a global graph using the connectivity relation from the neurite tree skeleton, the latent space velocity \(\tilde{\boldsymbol{u}}\), the distance embedding \(\boldsymbol{c}\) and the latent space concentration \(\tilde{z}\) is then assigned as the node features for the graph. Next,  the GCN block from the latent space system dynamic module \(\psi^{(z)}\), takes the graph as input to predict the time derivative of the latent space \(\frac{\partial \tilde{z}}{\partial t} \in \mathbb{R}^{T\times C\times 1\times1}\). This derivative is then used by a numerical integrator to calculate the state at the next time step, \(\tilde{z}(t_i+\Delta t)\). This operation is described as the following equation:
\begin{equation}
\begin{aligned}
        & \tilde{z}(t_i+\Delta t)=\text{Int}(\psi^{(n)}(\tilde{z}(t_i),\tilde{\boldsymbol{u}}, \boldsymbol{c}, t_i),\Delta t),\quad \text{for } t_i\in(0,T-\Delta t].\\
\end{aligned}
\end{equation}
The latent state \(\tilde{z}(t_i)\) is defined recursively. The process starts with the encoded initial condition and then uses the output from the previous integration step for all subsequent times.  This can be stated as:
\begin{equation}
\tilde{z}(t_i)=
\begin{cases}
    \bar{z}_0 & \text{if } t_i=0,\\
    \tilde{z}(t_i) & \text{if } t_i>0.
\end{cases}
\end{equation}
    \item \textbf{Step 4: 3D Concentration Reconstruction (Orange).} Finally, the predicted latent space concentration results on each time step (\(\tilde{z}=\{\tilde{n}_0, \tilde{n}_+\}\) where \(\tilde{n}_0, \tilde{n}_+\in \mathbb{R}^{T\times C\times1\times1}\)) are transformed back into the physical space using the concentration decoder \(\phi_{d}^{(n)}\):
    \begin{equation}
       \hat{z} = \{\hat{n}_0, \hat{n}_+\} = \phi_{d}^{(n)}(\tilde{n}_0, \tilde{n}_+, \boldsymbol{A}).
    \end{equation}
    
\end{itemize}
\section{Numerical testings and discussion} \label{ch5}
To evaluate the effectiveness and generalizability of the proposed GALDS framework, we present two case studies under distinct physical settings. The first case involves constant transport parameters \(k_+\) and \(k'_+\), which result in symmetric flow profiles. This case adopts the base GALDS architecture introduced in Section \ref{ch4} without further modifications. In contrast, the second case considers spatially-varying transport parameters, leading to non-symmetric flow patterns and increased modeling complexity. To accommodate this, the GALDS framework is extended with a pair of autoencoders and two additional input channels. The autoencoders serve two purposes: encoding the spatially varying parameters into a latent representation and enabling the model to reconstruct asymmetric flow features in the prediction stage. Despite these additions, the overall structure of GALDS remains largely unchanged, highlighting the adaptability of the framework to a range of transport conditions with minimal architectural overhead. For both cases, we analyze prediction accuracy, generalization to unseen geometries, and computational efficiency. 
\subsection{Case I: Normal material transport in complex neurite trees} \label{ch5.1}
In this case, we employ the base GALDS architecture to model symmetric flow profiles under constant transport parameters \(k_+\) and \(k'_+\). We focus on dataset generation, error comparison on testing and unseen geometries, and efficiency comparison.

\subsubsection{Dataset generation and testing error comparison} \label{ch5.1.1}
Our training dataset was generated using the IGA solver, which performs both Navier-Stokes and material transport simulations. These simulations were conducted on two distinct neuron geometries obtained from the NeuroMorpho.Org data base: NMO\_66731 and NMO\_66748. For each geometry, we executed 10 different boundary conditions while keeping the material transport parameters (as defined in Eq. (\ref{Eq: Material Transport})) constant at \(D=1.0\mu m^{2}/s\), \(k_{+}=1.0s^{-1}\), \(k'_{+}=0.5\) and the inlet velocity (as defined in Eq. (\ref{Eq: Navier-Stokes})) at \(u_{i}=0.1\mu m/s\), and the kinematic viscosity at \(\nu=0.1 \mu m^2/s\).  To ensure high accuracy, the time step for each simulation was set to  \(\Delta t=0.01s\). Simulations were run until a steady state was reached, at which point all spatial and temporal data were extracted. The collected simulation data were subsequently divided, with 80\% allocated for training and 20\% reserved for testing.

Table \ref{tab:Data generation cost} details the computational cost associated with generating a single simulation run (i.e., for one boundary condition) for each neuron geometry, along with the corresponding number of pipe and bifurcation elements. For model training, we adhered to the sequential procedure outlined in Algorithm \ref{alg:training} in Appendix. All training processes utilized full-batch optimization with the Adam optimizer, and the learning rate was set to \(10^{-4}\). The graph autoencoder model employs four hidden GConv layers and one feedforward layer. The encoder consists of four GConv layers followed by one feedforward layer, while the decoder begins with one feedforward layer followed by four GConv layers. The latent space system dynamics model, \(\psi_2\), comprised five hidden graph convolutional layers with a softplus activation function.

\begin{table}[h!]
\centering
\caption{Details of the generated dataset per single boundary condition. Our full dataset comprised 10 such boundary conditions for each model.}
\begin{tabular}{lcccc}
\hline
Model name & Number of elements & Pipe number & Bifurcation number & IGA computation cost \\
\hline
NMO\_66731 & 112,500 & 494 & 15 & 1,747 Core Hour\\
NMO\_66748 & 249,660 & 1,111 & 34 & 3,584 Core Hour\\
\hline
\end{tabular}
\label{tab:Data generation cost}
\end{table}

Following the training phase, each sub-model's performance was rigorously evaluated using its respective testing dataset. Table \ref{tab:Error evaluation} presents an error analysis on testing dataset for each module, demonstrating their performance on unseen data. The error metric used for this evaluation is the Mean Relative Error (MRE), calculated as:

\begin{equation}
    \text{MRE}=\frac{\sqrt{\sum_{i=1}^{N}{}(c^h_i-c^{\text{exact}}_i)^2}}{\text{max}||c^{\text{exact}}|| - \text{min}||c^{\text{exact}}||} \times 100\%,
\end{equation}

\noindent where \(N\) is the number of data points, \(c^h\) denotes model's prediction value, and \(c^{\text{exact}}\) is the corresponding exact value from the IGA simulation.

As shown in Table \ref{tab:Error evaluation}, both the velocity autoencoder (\(\phi^{(u)}\)) and the latent space transformation model (\(\psi^{(u)}\)) exhibit highly accurate reconstruction and prediction capabilities. This confirms their effectiveness in compressing physical information into the latent space and accurately mapping 1D insights, respectively. The concentration autoencoder (\(\phi^{(n)}\)) also demonstrates strong performance, with MREs of approximately \(1 \%\) for pipe geometries and below \(1 \%\) for bifurcation geometries, the Maximum Relative Error (MaxRE) also are also below \(2\%\). These low error values for individual components are critical, as they ensure that the downstream models in the GALDS pipeline receive high-fidelity latent space representations. The latent space system dynamics model, \(\psi^{(z)}\) is not included in this table because its training utilized data from the entire tree structure, making a direct separation into independent training and testing subsets for isolated evaluation challenging. Its performance will be assessed separately in Section \ref{ch5.1.2}.

\begin{table}[h!]
\centering
\caption{Reconstruction and prediction MRE and MaxRE evaluation on testing datasets for individual model components. The error values are formatted as MRE/MaxRE for each case.}
\begin{tabular}{lcc}
\hline
 & Pipe & Bifurcation \\
\hline
Velocity autoencoder \(\phi^{(u)}\)& \(<1\%/1.2\%\) & \(<1\%/1.6\%\)\\
Latent space transformation module \(\psi^{(u)}\)& \multicolumn{2}{c}{\(<1\%/1.6\%\)} \\
Concentration autoencoder \(\phi^{(n)}\)&  \(\approx1\%\)/1.5\%&\(<1\%/1.8\%\) \\
\hline
\end{tabular}
\label{tab:Error evaluation}
\vspace{0.3em}
\begin{minipage}{\linewidth} 
\raggedright\footnotesize
\textbf{Note:} Unlike other models in this table, the latent space transformation module presents only one pair of error values. This is due to its use of a single module to predict a unified latent  space for both pipe and bifurcation geometries (see Section \ref{ch4.2}).
\end{minipage}
\end{table}

\subsubsection{Error comparison on unseen geometries} \label{ch5.1.2}
This section evaluates the generalization capability of the entire GALDS model pipeline by assessing its performance on unseen neurite tree geometries. For this analysis, we selected eight distinct geometries: two from zebrafish neurons and six from mouse neurons. For each case, we first ran the full IGA simulation to obtain ground truth velocity and concentration fields. The GALDS predictions were then compared against these ground truth results to calculate the MRE and MaxRE. Table \ref{tab:Computational cost} summarizes these accuracy metrics and compares them against the PGNN method \cite{li2021deep}. While the table also presents computational cost, that aspect will be analyzed in detail in the following section; here, we will focus exclusively on the model's prediction accuracy.

\begin{table}[h!]
\centering
\caption{Comparison of prediction accuracy and computational cost between conventional IGA, PGNN and proposed GALDS on unseen geometries. Prediction accuracy is measured by MRE and MaxRE. For computational cost, IGA times refer to core minutes used in full numerical simulations, while PGNN and GALDS times indicate inference (evaluation) time.}
\label{tab:Computational cost}
\begin{tabular}{l|ccc|rcc}
\hline
 & \multicolumn{3}{c|}{MRE/MaxRE}&\multicolumn{3}{c}{Time (mins)}\\
\hline
Model Name & PGNN (\(n\)) & GALDS (\(n\)) & GALDS (\(\boldsymbol{u}\)) & IGA & PGNN & GALDS\\
\hline
NMO\_06840 &\(7.5\%\) & \(\boldsymbol{2.5\%/4.0\%}\) & \(0.1\%/2.0\%\) & 197,400 & 4.4 & 0.03\\
NMO\_06846 &\(7.2\%\) & \(\boldsymbol{2.7\%/5.3\%}\) & \(0.2\%/2.0\%\) & 92,520 & 2.1 & 0.06 \\
NMO\_32235 &\(7.8\%\) & \(\boldsymbol{1.5\%/5.0\%}\) & \(0.4\%/3.1\%\) & 53,760 & 1.3 & 0.10\\
NMO\_32280 &\(8.1\%\) & \(\boldsymbol{1.0\%/5.0\%}\) & \(0.4\%/3.9\%\)& 110,460 & 1.5 & 0.07\\
NMO\_54499 &\(8.7\%\) & \(\boldsymbol{2.4\%/3.4\%}\) & \(0.2\%/2.0\%\)& 425,040 & 6.1 & 0.09\\
NMO\_54504 &\(8.3\%\) & \(\boldsymbol{2.3\%/4.8\%}\) & \(0.3\%/1.8\%\)& 97,620 & 3.1 & 0.12\\
NMO\_112145 &\(7.4\%\) & \(\boldsymbol{1.3\%/3.1\%}\) & \(0.1\%/0.5\%\) & 78,420 & 1.2 & 0.11\\
NMO\_00865 &\(9.1\%\) & \(\boldsymbol{1.6\%/5.7\%}\) & \(0.3\%/5.7\%\) & 1,016,940 & 7.1 & 0.08\\
\hline
\end{tabular}
\vspace{0.3em}

\raggedright
{\footnotesize
\textbf{Note:} The PGNN results include only MRE, as MaxRE was not reported in the original paper \cite{li2021deep}.
\par}
\end{table}

\begin{figure}[htp]
\centering
\setlength{\abovecaptionskip}{5pt}
\setlength{\belowcaptionskip}{-10pt}
\includegraphics[width=1.0\linewidth]{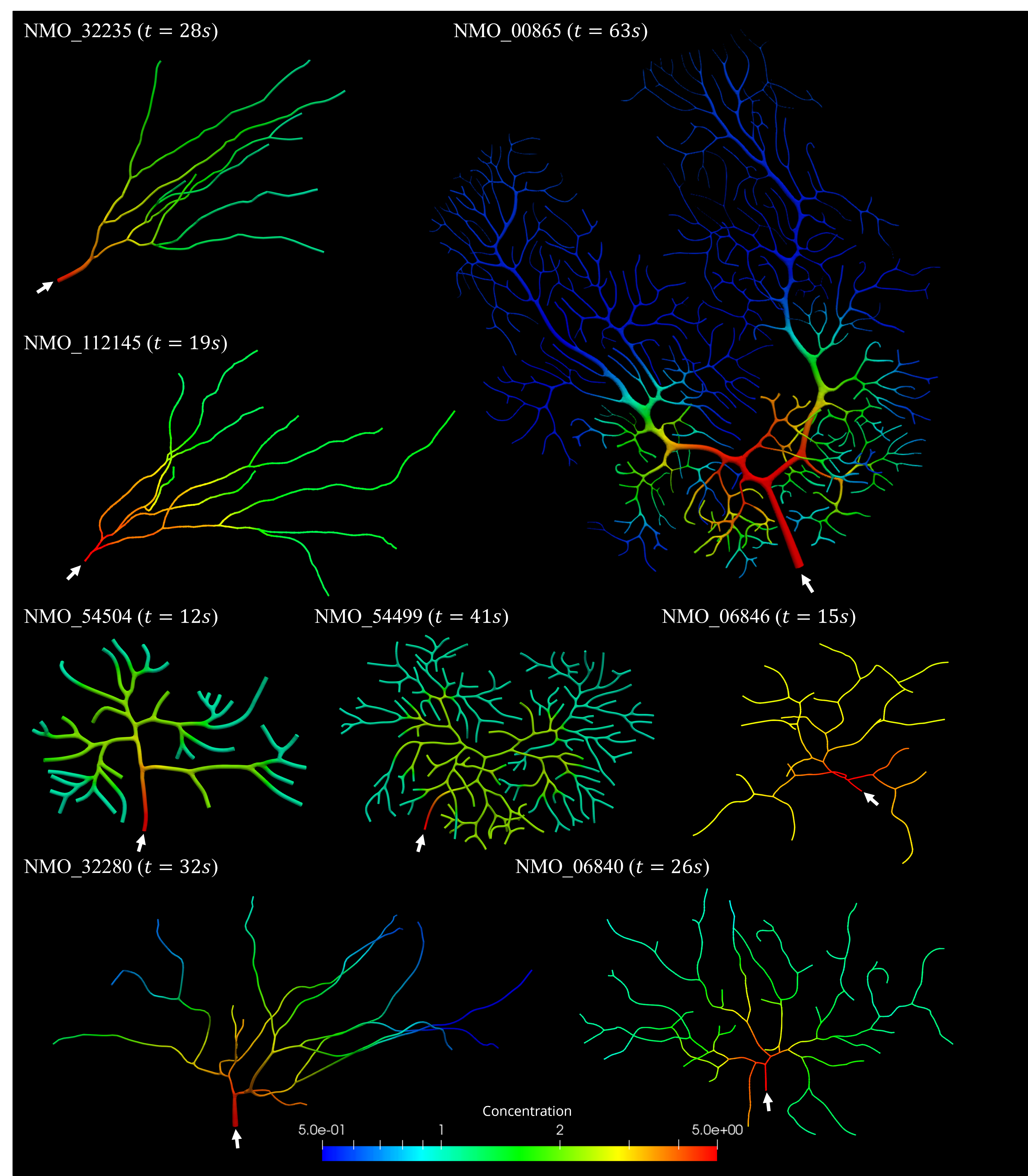}
\caption{Predicted concentration fields within the neurite network geometries for NMO\_06840, NMO\_06846, NMO\_32235, NMO\_32280, NMO\_54499, NMO\_54504, NMO\_112145, and NMO\_00865. Each result displays the steady-state prediction at the specific time point indicated in its label. The white arrows point to the inlet of the material.}
\label{fig:Whole pipeline prediction}
\end{figure}

\begin{figure}[htp]
\centering
\setlength{\abovecaptionskip}{5pt}
\setlength{\belowcaptionskip}{-10pt}
\includegraphics[width=1.0\linewidth]{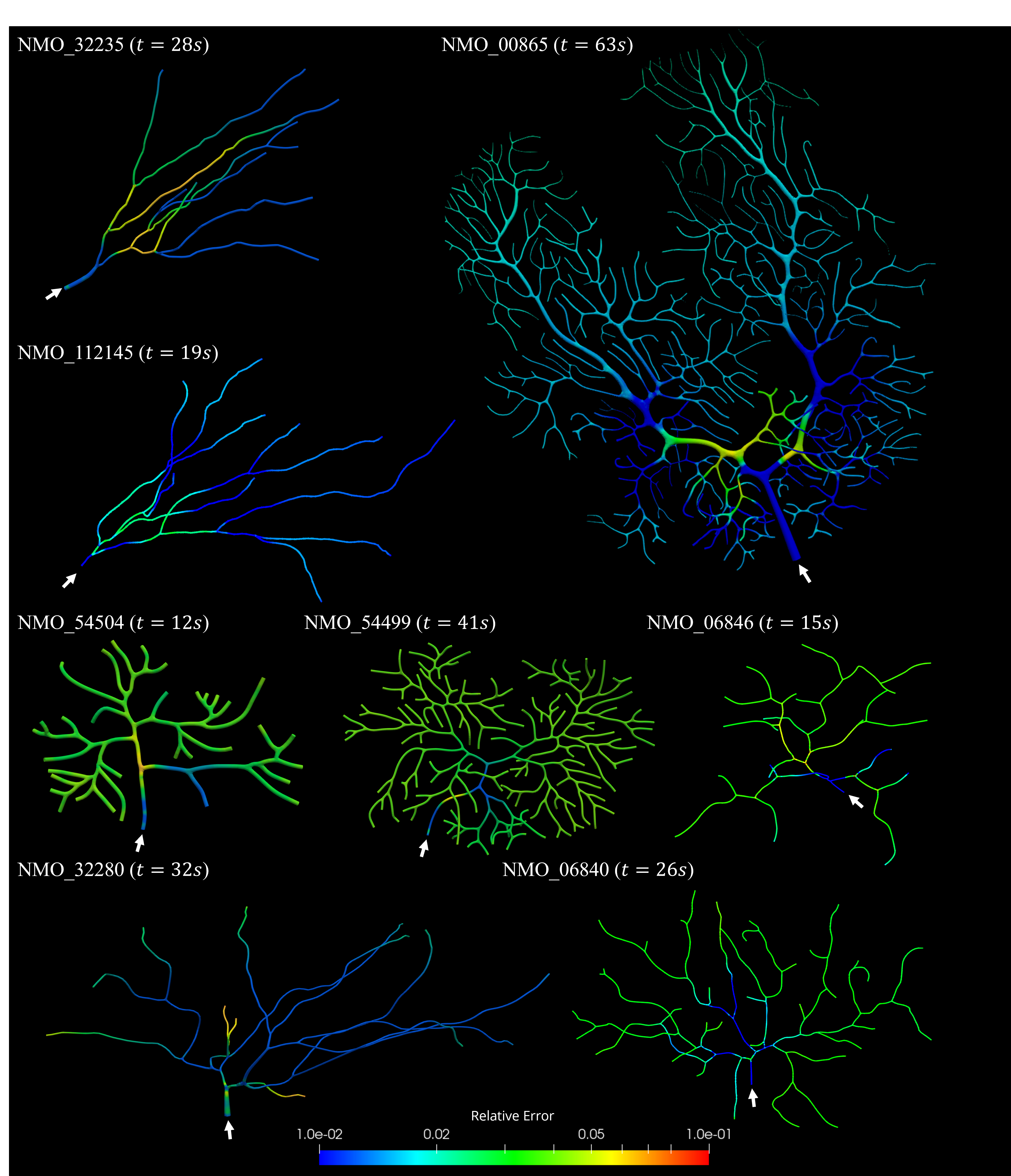}
\caption{Nodal relative error maps for concentration predictions within the neurite network geometries for NMO\_06840, NMO\_06846, NMO\_32235, NMO\_32280, NMO\_54499, NMO\_54504, NMO\_112145, and NMO\_00865. Each result displays the error distribution at the steady-state time point indicated in its label. The white arrows point to the inlet of the material. A single color bar is used across all results for clarity and the MRE and MaxRE are detailed in Table \ref{tab:Computational cost}.}
\label{fig:Whole pipeline prediction error}
\end{figure}

For the concentration field, GALDS achieves an MRE between 1.0\% and 2.7\%, a substantial improvement over the PGNN method's reported 7.2\% to 9.1\%. Crucially, GALDS also maintains a low MaxRE, peaking at 5.7\%. This metric is arguably more important for practical applications, as it bounds the worst-case error and ensures the model avoids catastrophic failures in local regions, suggesting a more reliable predictive tool. The qualitative predictions in Figure \ref{fig:Whole pipeline prediction} visually confirm these strong results. A deeper analysis of the error distribution in Figure \ref{fig:Whole pipeline prediction error} provides further insight, revealing that the largest concentration errors are localized in regions with steep concentration gradients. This is a common challenge for neural network surrogates, yet the low MaxRE values confirm that even in these regions, the model's predictions remain well within an acceptable error margin.

\begin{figure}[htb] 
\centering
\setlength{\abovecaptionskip}{5pt}
\setlength{\belowcaptionskip}{-10pt}
\includegraphics[width=1.0\linewidth]{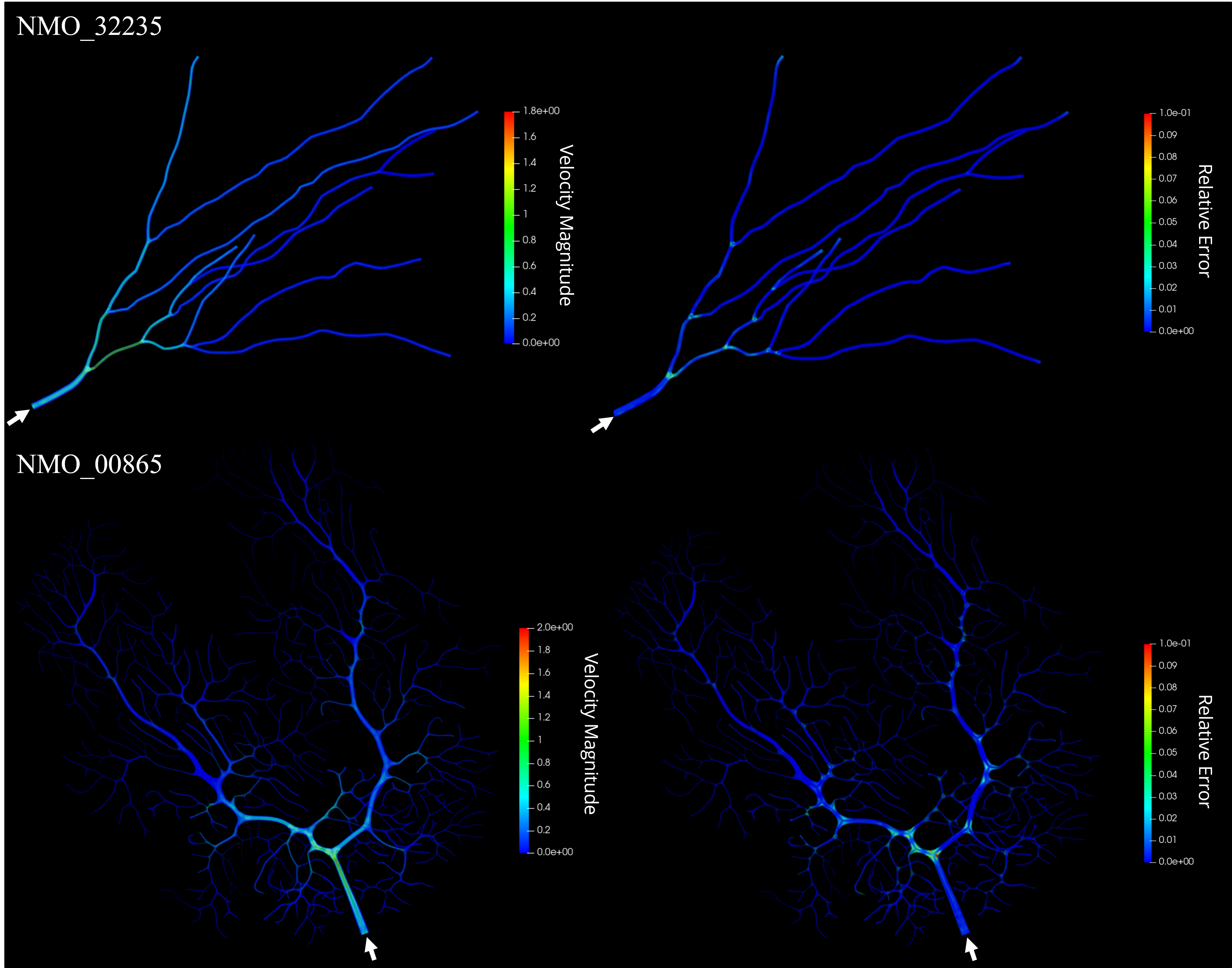}
\caption{Predicted velocity fields and corresponding relative error maps for two neurite geometries: NMO\_32235, NMO\_00865. Transparency is applied for clear visualization of the interior flow dynamics and white arrows point to the inlet.}
\label{fig:Velocity field relative error}
\end{figure}

The high accuracy in the concentration field is fundamentally dependent on the accuracy of the underlying velocity field, which dictates convective transport. As shown in Table \ref{tab:Computational cost}, the velocity predictions are exceptionally accurate, with an MRE consistently below 0.4\% and a MaxRE under 5.7\%. For a qualitative analysis of this, we selected two representative cases, NMO\_32235 and NMO\_00865, to visualize in Fig. \ref{fig:Velocity field relative error}. These visualizations confirm that the model correctly predicts the changes in the velocity field at bifurcations and in constricted passages, demonstrating a learned understanding of fundamental fluid dynamics. Interestingly, the velocity errors are most prominent at these bifurcation points. This observation points directly back to our training data generation (Table \ref{tab:Data generation cost}), where pipe sections constitute a much larger portion of the dataset than bifurcation sections. This data imbalance makes the complex junction flow patterns harder to learn and highlights a clear path for future improvement by enriching the training dataset. In summary, the high-fidelity velocity prediction serves as a solid foundation for the subsequent prediction of the concentration field. While minor inaccuracies exist and are traceable to specific, addressable challenges like data imbalance and steep physical gradients, the overall performance on unseen geometries confirms the powerful generalization capabilities of the GALDS pipeline.

\subsubsection{Efficiency comparison: Dataset, trainable parameters, and training time} \label{ch5.1.3}

The computational inference time comparison, detailed in Table \ref{tab:Computational cost}, highlights a significant efficiency gain. Each full IGA simulation is computationally expensive, whereas the evaluation process of our surrogate model yields predictions very rapidly. Specifically, our GLADS method can save over \(99\%\) of computational resources during inference compared to conventional IGA simulations, once the model is trained. When compared to PGNN, another surrogate model designed for efficiency, our proposed GALDS method achieves approximately a 10-fold speed-up in inference time, reducing prediction times from minutes to seconds.

Beyond inference speed, we also conducted a comprehensive efficiency comparison between GALDS and PGNN across three key dimensions: the size of the training dataset, the number of trainable parameters, and the total training time. The results are summarized in Table \ref{tab: Training time and model parameter comparison}. As evident from the table, GALDS demonstrates superior efficiency in all aspects: it requires a training dataset that is 20 times smaller, utilizes approximately 10 times fewer trainable parameters, and achieves a training speed that is six times faster.

As discussed in Section \ref{ch4}, the substantial decrease in required dataset size and the number of trainable parameters is primarily attributed to GALDS's approach of predicting results in a low-dimensional latent space. This information condensation is made possible by the integration of autoencoder architectures. The accelerated training speed is a direct consequence of both the fewer trainable parameters, which lead to reduced backpropagation computation, and the smaller training dataset, which results in faster iterations and data loading.

\begin{table}[h!]
\centering
\caption{Comparison of training efficiency between previous models (PGNN) and the proposed GALDS method, considering dataset size, trainable parameters, and training time.}
\begin{tabular}{lcrcc} 
\hline
Model & Model component & \shortstack{\rule{0pt}{11pt}Number of \\trainable parameters} & \shortstack{\rule{0pt}{11pt}Training dataset \\(IGA runs)} &  \shortstack{\rule{0pt}{11pt}Training time\\ (hrs)}\\
\hline
\multirow{4}{*}{PGNN} & Pipe simulator & \(728,000\) & \multirow{4}{*}{200} & \multirow{4}{*}{23.4} \\
                                        & Bifurcation simulator & \(728,000\) & & \\
                                        & Assembly model & \(48,000\) & & \\
\cline{2-3}
                                        & \rule{0pt}{10pt}Total & \rule{0pt}{10pt}1,504,000 & & \\
\hline
\multirow{5}{*}{GALDS}& Autoencoder \(\phi^{(u)}\)& 4,764 & \multirow{5}{*}{\(\boldsymbol{10}\)} & \multirow{5}{*}{\(\boldsymbol{4.7}\)} \\
                                & Latent space transformation model \(\psi^{(u)}\)& 33,923 &  &  \\
                                & Autoencoder \(\phi^{(n)}\)& 3,924 & & \\
                                & Latent space system dynamic model \(\psi^{(z)}\)& 83,842 & & \\
\cline{2-3}
                                & \rule{0pt}{10pt}Total & \rule{0pt}{10pt}\(\boldsymbol{126,453}\) & & \\
\hline
\end{tabular}
\label{tab: Training time and model parameter comparison}
\end{table}

\subsection{Case II: Traffic jam induced by spatially heterogeneous attachment kinetics } \label{ch5.2}
Intracellular transport is tightly regulated by motor protein attachment and detachment kinetics along cytoskeletal filaments. Real-world experiments have demonstrated that these kinetics vary spatially due to microtubule structural defects, such as protofilament number changes and tubulin subunit loss. For example, it has been shown that naturally occurring and engineered microtubule defects significantly disrupt the sustained movement of kinesin-1 motors  at specific sites, leading to disrupted processivity and localized motor accumulation \cite{gramlich2017single}. In addition, high-precision force-clamp experiments revealed that microtubules exhibit supertwist and lattice defects that further contribute to heterogeneous motor-filament interactions along the filament axis \cite{bugiel2018measuring}. These findings establish that spatial heterogeneity in motor attachment and detachment behavior is not only biologically plausible, but experimentally observed.

In this section, we present a numerical example designed to capture the impact of spatially varying attachment and detachment kinetics on transport dynamics. The first subsection details the necessary model adaptations and the generation of the simulation dataset used for training and validation. Next, we provide a quantitative comparison of prediction errors to assess the surrogate model’s accuracy in reproducing transport behavior under heterogeneous parameter conditions. Finally, an efficiency comparison evaluates the computational savings achieved by the surrogate model relative to conventional simulation approaches.

Here, we apply the GALDS framework to this scenario to accurately capture the resulting transport dynamics under spatially heterogeneous parameters. In this section, we detail the dataset generation, model modifications, prediction performance, and computational costs associated with this scenario.

\begin{figure}[htb]
\centering
\setlength{\abovecaptionskip}{-5pt}
\setlength{\belowcaptionskip}{-10pt}
\includegraphics[width=1.0\linewidth]{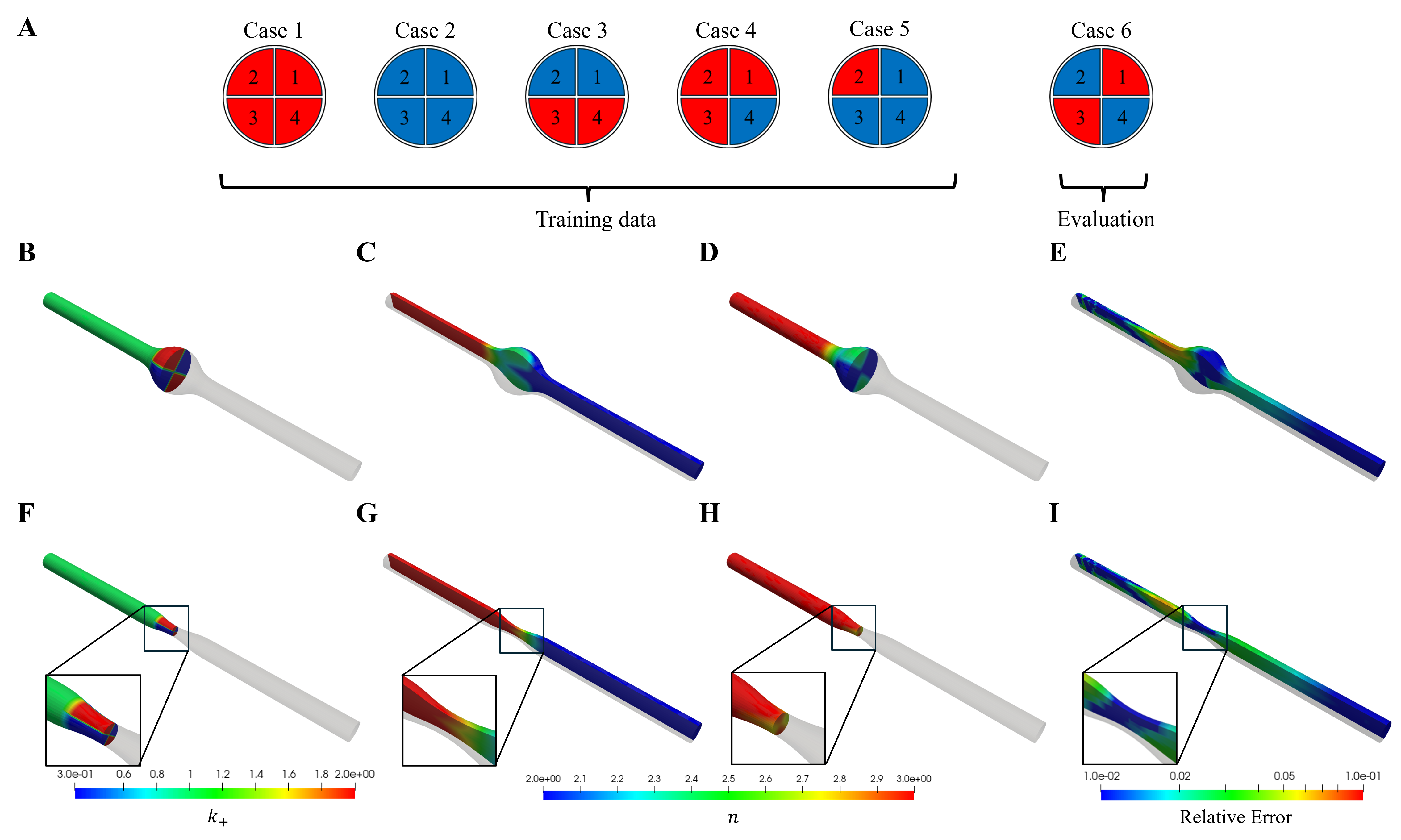}
\caption{Spatially asymmetric attachment rates and resulting flow patterns in swelling and shrinking pipes. (A) Five training and one evaluation cases of spatially asymmetric attachment rate \((k_+)\) profiles. Blue region represents low attachment rate \((k_+=0.3)\) and red region represents high attachment rate \((k_+=2.0)\). (B, F) Cross-sectional view of the attachment rate distribution for one case in swelling and shrinking pipes, with \(k_+=2.0\) for quadrants 1 and 3, and \(k_+=0.3\) for quadrants 2 and 4. (C, D, G, H) Two cross-secitonal views of the corresponding asymmetric concentration profile resulting from the transport simulation in the swelling and shrinking pipes. Note that the colorbar in (C, D, G, H) is set to range \([2.0, 3.0]\) to provide a clearer visualization of asymmetric concentration. (E, I) Relative error distribution of GALDS prediction in the swelling and shrinking pipes. A zoomed in view near the shrinking region is provided in (F-I) for better visualization.}
\label{fig:Assymetric pipe dome}
\end{figure}

\subsubsection{Model adaptation and dataset generation}

Our dataset was generated using an IGA solver on a single pipe geometry, meticulously exploring both swelling and shrinking conditions. We established three key attachment rates: high \((k_+=2.0)\), low \((k_+=0.3)\), and normal \((k_+=1.0)\). In our simulations, normal pipe regions were assigned the normal attachment rate. In contrast, abnormal (swelling or shrinking) regions were deliberately segmented into four quadrants, with each quadrant then assigned either a high or low attachment rate. This methodology allowed us to create six distinct asymmetric cases for each pipe geometry.

Fig. \ref{fig:Assymetric pipe dome}(A) presents six unique attachment rate profiles. While five of these cases were utilized for model training, the case exhibiting the most complex pattern, along with its IGA simulation result, was specifically reserved for evaluating the GALDS model's effectiveness. The impact of these spatial variations is further detailed in the figure: Figs. \ref{fig:Assymetric pipe dome}(B-D) illustrate the spatial distribution of \(k_+\) and two cross-sectional views of the resulting asymmetric flow patterns in the swelling geometry. Similarly, Figs. \ref{fig:Assymetric pipe dome}(F-H) display the corresponding parameter fields and flow behavior under shrinking deformation. The chosen patterns highlight how localized changes in transport parameters directly induce asymmetric flow. Consistent with Section \ref{ch5.1}'s data splitting strategy, 80\% of the simulation results were allocated for training, and 20\% for testing.

The choice to simulate both swelling and shrinking geometries is motivated by their relevance to neurite dysfunction and cytoskeletal heterogeneity in biological systems. Swelling occurs when neurites experience volume expansion under osmotic stress. This process is mechanically opposed by microtubules, which slow the rate of swelling and help regulate the onset of beading transitions in PC12 neurites \cite{fernandez2010role, datar2019roles}. In contrast, shrinking refers to a reduction in diameter that serves as an active protective response after axonal transsection. It is driven by cytoskeletal disassembly and water extrusion through aquaporins \cite{aydin2023active}. These two geometries represent biologically meaningful scenarios of structural change and allow us to capture spatial heterogeneity in motor protein attachment and detachment behavior, providing a foundation for surrogate modeling of asymmetric flow responses.

To handle the additional complexity introduced by the spatially-varying parameters, we extend the base GALDS architecture with two key modifications. First, two pairs of autoencoders are introduced. One pair is to encode the spatial distribution of transport parameters into a latent representation that can be processed by the latent space system dynamic model. It can be written as:

\begin{equation} \label{Eq: Auto-encoder model general for attachment rate}
\left \{
\begin{aligned}
    & \tilde{k} = \phi_{e}^{({k})}(k, \boldsymbol{A}; \theta_{e}^{({k})}),\\
    & \hat{k} = \phi_{d}^{({k})}(\tilde{k},\boldsymbol{A};\theta_{d}^{({k})}),\\
\end{aligned}
\right.
\end{equation}

\noindent where \(k\in \{k_+, k'_+\}\) represents that the two parameters share the same autoencoder pair. Another pair of autoencoder aims to encode and decode the asymmetric output features with the same  architecture as \(\phi^{(n)}\) in Section \ref{ch4.1} but trained with asymmetric profile results. 

These modules enhance the GALDS' ability to capture localized variations and reconstruct complex flow patterns. In addition, two additional input channels (\(k_+\) and \(k'_+\)) are added to the latent space system dynamic module to provide the information of the spatially-varying parameters. We have:

\begin{equation} \label{Eq: Nerual ODE asymmetric}
  \frac{\partial \tilde{z}}{\partial t} = \psi^{(z)}(\bar{z}_0, \bar{\boldsymbol{u}},\boldsymbol{c}, \bar{k}_+, \bar{k}'_+, t;\theta _{\psi}^{(z)}),
\end{equation}

\noindent where \(\bar{k}_+\) and \(\bar{k}'_+\) are the latent space representation of the two parameters given by the trained encoder \(\phi^{(k)}_e\).

Despite these additions, the overall structure of the GALDS framework remains largely intact, with the encoder-decoder backbone, latent space representation, Neural ODE system dynamics prediction pipeline preserved. The concentration decoder \(\phi^{(n)}_d\) , transport parameter autoencoder \(\phi^{(k)}\) and the latent space system dynamics module \(\psi^{(z)}\) need to be trained on the new dataset. Other components, including the velocity autoencoder \(\phi^{(u)}\) and the latent space transformation model \(\psi^{(u)}\), can directly reuse the trained parameters from Case I without any retraining.

\subsubsection{Error comparison and efficiency study}
\begin{figure}[htb]
\centering
\setlength{\abovecaptionskip}{-5pt}
\includegraphics[width=1.0\linewidth]{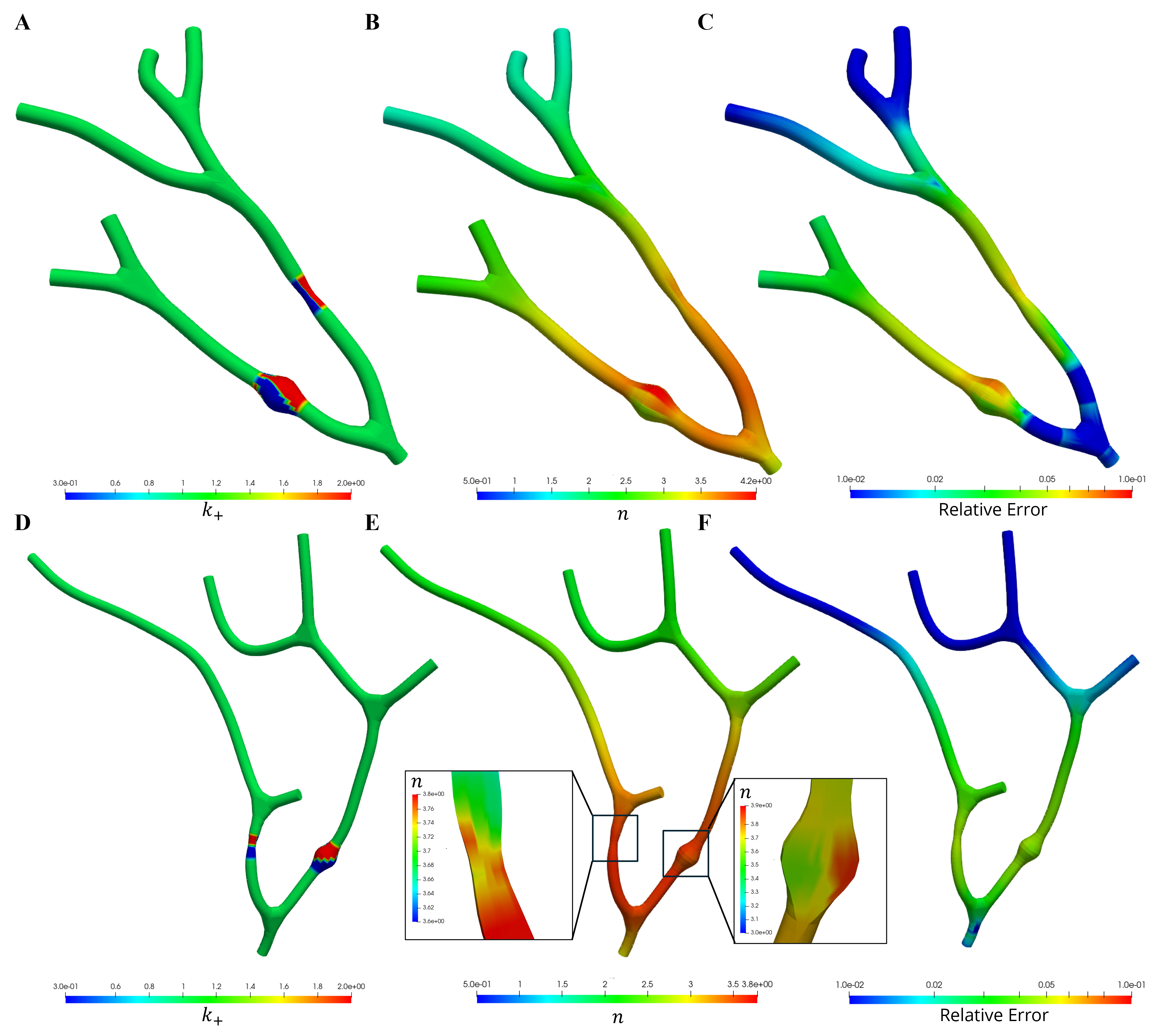}
\caption{GALDS prediction accuracy for spatially asymmetric attachment rate examples. (A-C) Results for the NMO\_54499 neurite subtree, showing the prescribed attachment rate profile (A), the GALDS-predicted concentration (B), and the relative error (C). (D-F) Corresponding results for the NMO\_66748 neurite subtree. The insets in (E) show two zoomed-in cross-sections detailing the asymmetric concentration. The model accurately captures concentration accumulation and complex profiles, with errors localized to high-gradient regions.}
\label{fig:case2 result}
\end{figure}

This section evaluates the GALDS framework's accuracy and efficiency on two types of challenging numerical examples. The first involves simple pipe geometries with swelling or shrinking profiles and previously unseen spatial distributions of transport parameters (as in Case 6, Fig. \ref{fig:Assymetric pipe dome}(A)). The second uses complex neurite tree geometries (NMO\_54499 and NMO\_66748 from NeuroMorpho.Org, shown in Fig. \ref{fig:case2 result}) with localized swelling/shrinking and spatially varying parameters. For each case, we generate ground trouth velocity and concentration fields using IGA simulations. We then compare the GALDS predictions to these references, quantifying performance with MRE and MaxRE. Since the velocity autoencoder and latent space transformation models were already validated in Section \ref{ch5.1.2}, we focus here on visualizing the concentration results and analyzing computational efficiency.

For the simple swelling and shrinking pipe geometries, the relative error is highest in the region immediately preceding the geometric alteration as shown in Fig. \ref{fig:Assymetric pipe dome}(E, I). This localization corresponds to a ``traffic jam" effect where the low attachment rate causes flow obstruction and a steep concentration gradient. In the more complex neurite geometries, GALDS demonstrates strong predictive capability. The results for the NMO\_54499 subtree (Fig. \ref{fig:case2 result}(A-C)) and the NMO\_66748 subtree (Fig. \ref{fig:case2 result}(D-F)) show that the model accurately captures key physical phenomena. Specifically, the predicted concentrations in Fig. \ref{fig:case2 result}(B, E) correctly show accumulation in regions with high attachment rates. Furthermore, GALDS successfully reproduces the asymmetric concentration profiles, a direct result of the fine-tuning applied to the concentration decdoer \(\phi^{(n)}_d\). As with previous examples, the relative error is primary concentrated in regions with high concentration gradients.

Table \ref{tab:computational_cost_jam} summarizes both the performance and efficiency of the GALDS framework. The velocity predictions remain highly accurate, with a MaxRE below \(4\%\), while the concentration predictions also demonstrate strong robustness, achieving an MRE of less than \(3\%\).The MaxRE for the concentration is relatively higher, ranging from \(6\%\) to \(8\%\), with the highest errors observed near the challenging regions exhibiting geometric or parametric abnormalities. These areas may benefit from further targeted training. In terms of computational efficiency, the inference time remains under one minute for three of the four test cases, and the model architecture ensures minimal training time. This substantial acceleration compared to conventional IGA simulations, while maintaining reasonable accuracy, enables large-scale parameter studies that would otherwise be computationally infeasible.

\begin{table}[h!]
\centering
\caption{GALDS prediction accuracy and computational cost comparison between IGA and GLADS on unseen spatial distributions and geometries. Accuracy is measured by MRE and MaxRE. Time is reported in minutes, with IGA times refurring to full numerical simulations and GALDS times separated into training and inference durations.}
\label{tab:computational_cost_jam}
\begin{tabular}{l|cc|rcc} 
\hline
 & \multicolumn{2}{c|}{MRE / MaxRE} & \multicolumn{3}{c}{Time (mins)} \\ 
\hline
Model Name & GALDS (\(n\)) & GALDS (\(\boldsymbol{u}\)) & \multicolumn{1}{c}{IGA}  & GALDS (Training) & GALDS (Inference) \\ 
\hline
Pipe\_swell (Case 6) & 1.9\%/7.8\%  & 0.3\%/2.2\%& 6,475 & \multirow{4}{*}{557}& 0.33\\ 
Pipe\_shrink (Case 6) & 2.0\%/6.9\% & 0.2\%/1.1\% & 6,461 & & 0.35\\ 
NMO\_54499\_SubTree & 2.7\%/7.5\% & 0.6\%/3.2\% & 19,336 & & 0.73\\ 
NMO\_66748\_SubTree & 2.6\%/7.8\% & 0.4\%/2.6\% & 53,224 & & 1.11\\ 
\hline
\end{tabular}
\vspace{0.3em}
\raggedright
\end{table}

\section{Conclusion and future work} \label{ch6}
In this work, we successfully developed the GALDS model, a novel framework designed for efficient and accurate simulation of neurite transport. At its core, GALDS leverages a specialized graph autoencoder architecture to compress high-dimensional physical simulation data (velocity and concentration fields) into a compact, low-dimension latent space. This approach is instrumental in decreasing the overall model size and significantly reducing the amount of training data required, compared to methods that operate directly in high-dimensional physical space.

A key innovation of our model lies in its utilization of Graph Neural ODEs to learn the complex system dynamics within such condensed latent space. By modeling the continuous time derivatives rather than discrete state transitions, our Graph Neural ODE component effectively mitigates the accumulated error inherent in traditional recurrent neural networks, leading to improved stability and generalization. Furthermore, our design judiciously incorporates physically-informed inputs, such as 1D Navier-Stokes simulation results, into the latent space transformation model, thereby enriching the model's predictive capability while keeping input requirements minimal.

The robustness and generalization capabilities of GALDS were rigorously validated through extensive numerical testing. We demonstrated its ability to accurately predict velocity and concentration fields on eight diverse, unseen neurite tree geometries (two zebrafish and six mouse neurons). Our error analysis confirmed high prediction accuracy, with mean relative errors consistently low and maximum relative errors well within acceptable ranges, as detailed in our comparative tables. Critically, GALDS offers substantial computational efficiencies: it reduces the required training dataset size by a factor of 20, utilizes approximately 10 times fewer trainable parameters, accelerates training time by six-fold, and provides a remarkable 10-fold speed-up in inference time compared to the state-of-the-art PGNN method. This superior efficiency is achieved by operating within the compressed latent space and by the streamlined integration of 1D simulation data. We then extended our validation to more complex scenarios featuring spatially varying transport parameters and geometric alterations designed to induce transport obstruction. On both simple pipe and complex neurite geometries, GALDS successfully captured challenging physical phenomena, including asymmetric concentration profiles with localized ``traffic jam" effects, demonstrating its robustness beyond baseline conditions. While MRE remained low (\(<3\%\)), these complex cases produced higher MaxRE (up to \(8\%\)), pinpointing the precise regions where model performance could be further enhanced.

Our proposed GALDS method offers a new and promising perspective on integrating GNNs for complex biological transport problems. It elegantly combines the ability to model local features at cross sections (through the graph autoencoders) with the capacity to capture global dynamics across entire tree-like structures (via the graph Neural ODEs). The dramatic speed-up in surrogate model prediction capabilities opens up exciting avenues for computationally intensive studies, such as comprehensive uncertainty quantification, extensive parameter studies in neurite transport simulations, and broader applications to other intricate tree-like structures, including vascular blood flow networks and root systems in plants.

It is important to note that all simulations in this study were conducted under laminar flow conditions, where GALDS performs robustly. However, looking toward broader applications, the model has key architectural limitations. Its reliance on an autoencoder, while efficient for global compression, tends to smooth over the fine-grained details necessary to resolve highly localized phenomena, which would be a challenge in potential applications involving turbulent flow. Furthermore, the framework is designed for static geometries and is currently unable to predict dynamic changes to the underlying structure, such as the growth or retraction of a neurite tree over time.

Future work will focus on addressing these limitations by exploring several directions. One primary avenue involves the development of specialized machine learning models or hybrid architectures specifically designed to predict and resolve these local features. This could include incorporating multi-resolution autoencoders, attention mechanisms to highlight critical local regions, or integrating PINNs that enforce local conservation laws more strictly. Furthermore, we aim to extend GALDS to incorporate more complex biophysical phenomena, such as dynamic changes in geometry, multi-species transport with more intricate reaction kinetics, or the ability to dynamically adapt the graph structure itself to capture evolving geometries. Long-term goals also include the integration of real-world experimental data to further validate and refine the model, alongside continued efforts to enhance model interpret ability and provide robust uncertainty quantification directly within the latent space framework.

\section{Appendix}
This section provides a detailed algorithmic description of the training process for the proposed GALDS model, structured into four distinct phases. Phase 1 details the training of the velocity autoencoders, and Phase 2 outlines the training of the latent space transformation model. Phase 3 then introduces the training of concentration autoencoders. Finally, Phase 4 describes the training of the latent space dynamics models.

\begin{algorithm}
\caption{Training Procedure for GALDS}\label{alg:training}
\begin{algorithmic}[1]

\Require Training dataset $D = \{(x_i, \boldsymbol{u}_i, n_{0,i}, n_{+,i})\}_{i=1}^N$
\Ensure Trained parameters $\theta_{\phi}^{(u)}, \theta_{\psi}^{(u)}, \theta_{\phi}^{(n)}, \theta_{\psi}^{(n)}$

\Statex \rule{\linewidth}{0.4pt}
\Statex \textbf{Phase 1: Train Velocity Autoencoders ($\phi_{e}^{(u)}, \phi_{d}^{(u)}$)}
\For{model type in \{\texttt{pipe}, \texttt{bifurcation}\}} \hfill \texttt{(in parallel)}
    \For{each epoch}
        \State $\boldsymbol{z}_{u} \gets \phi_{e}^{(u),model}(\boldsymbol{u}^{model}; \theta_{e}^{(u),model})$
        \State $\hat{\boldsymbol{u}} \gets \phi_{d}^{(u),model}(\boldsymbol{z}_{u}; \theta_{d}^{(u),model})$
        \State $\mathcal{L}_{\phi^{(u)}}^{model} \gets \|\hat{\boldsymbol{u}} - \boldsymbol{u}^{model}\|_2^2$
        \State Update $\theta_{e}^{(u),model}, \theta_{d}^{(u),model}$ to minimize $\mathcal{L}_{\phi^{(u)}}^{model}$
    \EndFor
\EndFor

\Statex \rule{\linewidth}{0.4pt}
\Statex \textbf{Phase 2: Train Latent Space Transformation Model ($\psi^{(u)}$)}
\For{each epoch}
    \State $\tilde{\boldsymbol{u}} \gets \psi^{(u)}(\boldsymbol{u}_{1D}; \theta_{\psi}^{(u)}); \quad\bar{\boldsymbol{u}} \gets \phi_{e}^{(u)}(\boldsymbol{u}_{3D}; \theta_{e}^{(u)}) $
    \State $\mathcal{L}_{\psi^{(u)}} \gets \|\tilde{\boldsymbol{u}} - \bar{\boldsymbol{u}}\|_2^2$
    \State Update $\theta_{\psi}^{(u)}$ to minimize $\mathcal{L}_{\psi^{(u)}}$
\EndFor

\Statex \rule{\linewidth}{0.4pt}
\Statex \textbf{Phase 3: Train Concentration Autoencoders ($\phi_{e}^{(n)}, \phi_{d}^{(n)}$)}
\For{model type in \{\texttt{pipe}, \texttt{bifurcation}\}} \hfill \texttt{(in parallel)}
    \For{each epoch}
        \State $(\bar{n}_0, \bar{n}_+) \gets \phi_{e}^{(n),model}(n_0^{model}, n_+^{model}; \theta_{e}^{(n),model})$
        \State $(\hat{n}_0, \hat{n}_+) \gets \phi_{d}^{(n),model}(\bar{n}_0, \bar{n}_+; \theta_{d}^{(n),model})$
        \State $\mathcal{L}_{\phi^{(n)}}^{model} \gets \|\hat{n}_0 - n_0^{model}\|_2^2 + \|\hat{n}_+ - n_+^{model}\|_2^2$
        \State Update $\theta_{e}^{(n),model}, \theta_{d}^{(n),model}$ to minimize $\mathcal{L}_{\phi^{(n)}}^{model}$
    \EndFor
\EndFor

\Statex \rule{\linewidth}{0.4pt}
\Statex \textbf{Phase 4: Train Latent Space Dynamics Models ($\psi^{(z)}$)}

\Statex \textbf{Step 1: Train models $\psi^{(z=n_0)}$ and $\psi^{(z=n_+)}$ separately} 
\For {target field $z\in[n_0, n_+]$}
    \For{each epoch}
        \State $\bar{z}_0 \gets \phi_{e}^{(n)}(z(t=0), \boldsymbol{A};\theta^{(n)}_e)$
        \State $\bar{\boldsymbol{u}} \gets \phi^{(u)}_e(\boldsymbol{u}, \boldsymbol{A};\theta_e^{(u)})$
        \State $\frac{\partial \boldsymbol{z}}{\partial t} \gets \psi^{(z)}(\bar{z}_0, \bar{\boldsymbol{u}}, t; \theta_{\psi}^{(z)})$
        \State $\tilde{z} \gets \text{Int}(\frac{\partial \boldsymbol{z}}{\partial t}, \Delta t)$
        \State $\mathcal{L}_{\psi^{(z)}} \gets \|\tilde{z} - \phi_{e}^{(n)}(z,\boldsymbol{A};\theta_e^{(n)})\|_2^2$
        \State Update $\theta_{\psi}^{(z)}$ to minimize $\mathcal{L}_{\psi^{(z)}}$
    \EndFor
\EndFor

\Statex \textbf{Step 2: Joint fine-tuning of $\psi^{(n_0)}$ and $\psi^{(n_+)}$}
\For {each epoch}
    \State $\tilde{n}_{\text{joint}} \gets \tilde{n}_0 + \tilde{n}_+ ; \quad n_{\text{joint}} \gets n_0 + n_+$
    \State $\mathcal{L}_{\psi^{(n_0+n_+)}} \gets \|\tilde{n}_{\text{joint}}-n_{\text{joint}}\|_2^2$
    \State Update $\theta_{\psi}^{(n_0)}$ and $\theta_{\psi}^{(n_+)}$ to minimize $\mathcal{L}_{\psi^{(n_0+n_+)}}$
\EndFor

\end{algorithmic}
\end{algorithm}


\section*{Code and data availability}
The codes in this paper are accessible in the "NeuronTransportGALDS" GitHub repository: \url{https://github.com/CMU-CBML/NeuronTransportGALDS}. Correspondence and requests for codes and data should be addressed to T. Y. Hsieh and Y. J. Zhang.

\section*{Declaration of competing interest}
The authors declare no known competing financial interests or personal relationships that could have appeared to influence the work reported in this paper.

\section*{Acknowledgement}
T. Y. Hsieh and Y. J. Zhang were supported by the US National Science Foundation grants CMMI-1953323 and CBET-2332084. This work used RM-nodes on Bridges-2 Supercomputer at Pittsburgh Supercomputer center through allocation ID eng170006p from the Advanced Cyberinfrastructure Coordination Ecosystem: Services \& Support (ACCESS) program, which is supported by the National Science Foundation, United States grants \#2138259, \#2138307, \#2137603, and \#2138296.

\bibliographystyle{elsarticle-num}
\bibliography{reference}
\end{document}